\documentclass{article}

\usepackage[preprint]{corl_2026} 

\usepackage{pifont}
\usepackage{graphicx}
\usepackage{amsmath,amssymb} 
\usepackage{booktabs}
\usepackage{caption}
\usepackage{makecell}

\usepackage{float}

\usepackage{pifont}
\usepackage{wrapfig}
\usepackage{multirow}
\usepackage{wrapfig}

\usepackage[ruled,vlined]{algorithm2e}
\SetKwInput{KwInput}{Input}
\SetKwInput{KwOutput}{Output}

\usepackage[accsupp]{axessibility}  
\usepackage{subcaption}
\usepackage{titletoc} 

\usepackage{caption}
\usepackage{color}
\usepackage[normalem]{ulem}
\definecolor{barrier}{RGB}{112,128,144}
\definecolor{bicycle}{RGB}{220,20,60}
\definecolor{bus}{RGB}{255, 127, 80}
\definecolor{car}{RGB}{255, 158, 0}
\definecolor{const. veh.}{RGB}{233, 150, 70}
\definecolor{motorcycle}{RGB}{255,61,99}
\definecolor{pedestrian}{RGB}{0,0,230}
\definecolor{traffic cone}{RGB}{47,79,79}
\definecolor{trailer}{RGB}{255,140,0}
\definecolor{truck}{RGB}{255,99,71}
\definecolor{drive. suf.}{RGB}{0,207,191}
\definecolor{other flat}{RGB}{175,0,75}
\definecolor{sidewalk}{RGB}{75,0,75}
\definecolor{terrain}{RGB}{112,180,60}
\definecolor{manmade}{RGB}{222,184,135}
\definecolor{vegetation}{RGB}{0,175,0}

\definecolor{background}{RGB}{128,128,128}

\definecolor{road}{RGB}{31,119,180}

\definecolor{sidewalk}{RGB}{174,199,232}

\definecolor{crosswalk}{RGB}{255,127,14}

\definecolor{grass}{RGB}{255,187,120}

\definecolor{wall}{RGB}{44,160,44}

\definecolor{trees}{RGB}{152,223,138}

\definecolor{vehicle}{RGB}{214,39,40}

\definecolor{cyclist}{RGB}{255,152,150}

\definecolor{animal}{RGB}{148,103,189}

\definecolor{generic_obstacle}{RGB}{197,176,213}

\definecolor{pedestrian}{RGB}{140,86,75}

\definecolor{driveway}{RGB}{196,156,148}

\definecolor{curb}{RGB}{227,119,194}

\definecolor{gutter}{RGB}{247,182,210}

\definecolor{COCOa1}{RGB}{255,120,80}
\definecolor{COCOa2}{RGB}{255,100,100}
\definecolor{COCOa3}{RGB}{235,80,120}
\definecolor{COCOa4}{RGB}{215,60,140}
\definecolor{COCOa5}{RGB}{195,40,160}

\newcommand{\DatasetName}{Sidewalk3D}
\newcommand{\ModelName}{WalkOCC}

\usepackage{hyperref}

\title{
Monocular 3D Occupancy Perception for Robots on Sidewalks via Hybrid 2D--3D Learning
}

\author{
    Yukai Ma$^{1}$
    \quad
    Joe Lin$^{2}$
    \quad
    Liu Liu$^{1,3}$
    \quad
    Honglin He$^{1}$
    \quad
    Lulu Ricketts$^{2}$ 
    \\
    \textbf{
    Brad Squicciarini$^{2}$ 
    \quad
    Bolei Zhou$^{1,2}$ 
    }
    \\
    $^{1}$ University of California, Los Angeles
    \quad \\
    $^{2}$ Coco Robotics \quad
    $^{3}$ Massachusetts Institute of Technology\\
    \textbf{\url{https://vail-ucla.github.io/walkocc/}}
}

\begin{document}
\maketitle


\begin{abstract}
Sidewalks in the real world are crowded, cluttered, and less structured than roads, making 3D occupancy prediction a key ingredient for the safe navigation of mobile robots such as delivery bots and electric wheelchairs. Existing occupancy learning pipelines are largely designed for on-road autonomous driving and often train on large-scale paired LiDAR-RGB datasets with dense 3D supervision and multiple camera inputs, which are costly to collect and do not adequately capture sidewalk-specific characteristics.
We propose {\ModelName}, a hybrid Ray-marching monocular 3D occupancy perception framework for robots operating on sidewalks. {\ModelName} explicitly couples geometric grounding from LiDAR-RGB paired data with scalable learning from large-scale unpaired monocular images. It bootstraps pseudo occupancy supervision from paired sequences and jointly learns image-level representations on additional 2D-only data. It yields stable optimization and improved generalization without requiring costly 3D occupancy annotations.
Extensive experiments demonstrate consistent gains in prediction accuracy, fine-grained segmentation of subtle urban structures such as curbs and gutters, and robustness to environmental and cross-embodiment shifts compared with self-supervised image-based baselines. To facilitate evaluation and benchmarking, we also introduce \DatasetName, a large-scale sidewalk perception dataset with LiDAR-camera paired sequences collected across multiple locations and time periods, along with 3D semantic occupancy annotations for evaluation. Code and data will be made available.
\end{abstract}

\keywords{Robot Perception, Sidewalks Occupancy Prediction, Weak Supervision}


\section{Introduction}

\label{sec:intro}

\begin{figure}[t]
    \centering
    \includegraphics[width=\linewidth]{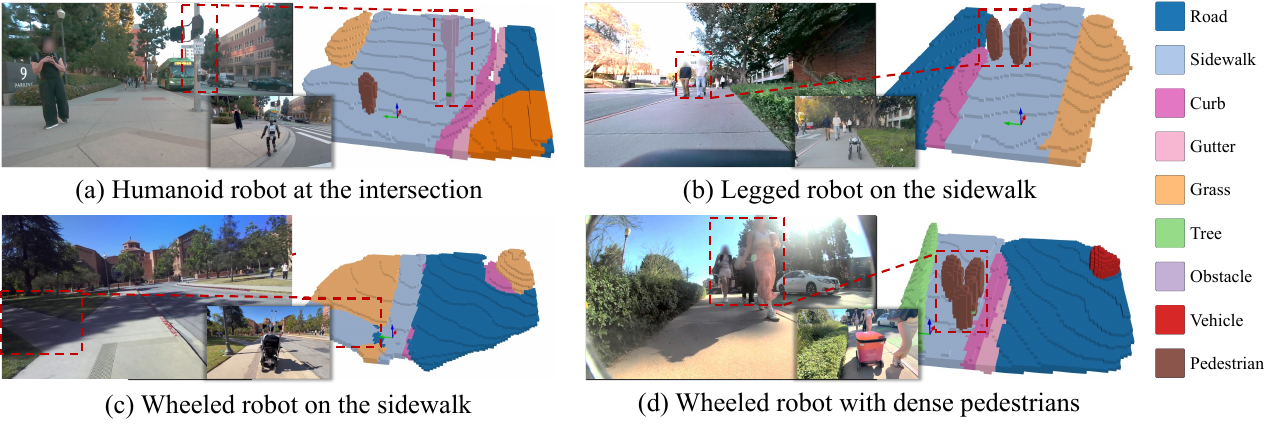}
    \caption{
    \textbf{3D occupancy prediction of challenging real-world sidewalk scenes for various mobile robots.} We plot four prediction results achieved by our {\ModelName} for four different robot embodiments, including a humanoid robot, a quadruped robot, an electric wheelchair, and a wheeled delivery robot. Each example shows the third-person-view image, the predicted occupancy output, and the model input image in the corner for reference. The color legend for semantic categories is plotted on the right. The coordinate axes in the occupancy prediction visualization indicate the robot's current camera pose in a right-handed coordinate system, with red, green, and blue corresponding to the $x$, $y$, and $z$ axes, respectively. 
    }
    \label{fig:teaser}
\end{figure}

Mobile robots are increasingly being deployed on urban sidewalks for applications such as food delivery and personal mobility assistance. Unlike on-road autonomous driving, sidewalk environments are less structured and more cluttered, with dense pedestrian traffic and diverse obstacles. In this setting, robust situational awareness is essential for safe navigation. In autonomous driving, 3D semantic occupancy prediction~\cite{yu2023flashocc,tian2023occ3d,wei2023surroundocc,ma2024licrocc,wang2025l2cocc}, which estimates whether surrounding 3D voxels are occupied and assigns semantic labels (e.g., car, road, people), has proven effective for scene understanding and downstream planning. However, sidewalk occupancy perception for mobile robots remains much less explored, as shown in Figure~\ref{fig:teaser}.

Most existing 3D occupancy pipelines are built for autonomous driving and rely on synchronized image-LiDAR pairs. Supervised methods such as SurroundOcc~\cite{wei2023surroundocc}, OCC3D~\cite{tian2023occ3d}, and OpenOccupancy~\cite{wang2023openoccupancy} require dense 3D supervision derived from semantically labeled point clouds, which is expensive to obtain. More label-efficient alternatives, \textit{e.g.}, RenderOcc~\cite{pan2024renderocc} and SceneDINO~\cite{jevtic2025feed}, reduce manual labeling but still depend on high-fidelity 3D reconstruction. For sidewalk data captured from a single moving viewpoint with limited multi-view consistency, such reconstruction is often unreliable, which makes rendering-based supervision unstable.
Existing human-centric urban robotics datasets~\cite{karnan2022socially,martin2019jrdb,carlevaris2016university} employ sensors and viewpoints similar to sidewalk robots, but they are often collected on campuses or in controlled areas and exhibit a notable domain gap relative to real-world sidewalk deployment. 

Scaling RGB-LiDAR paired data remains a bottleneck because it requires multi-sensor hardware and accurate per-robot extrinsic calibration, limiting collection scale and diversity. Generalization is further challenged by intrinsic discrepancies across robot platforms, including different sensor layouts and body morphologies, in addition to external shifts such as region and illumination changes. In contrast, monocular images are abundant and easy to acquire, offering broader scene coverage while naturally accommodating diverse robot embodiments. Motivated by this, we study monocular 3D occupancy learning that leverages unpaired 2D images to improve generalization across both environmental and cross-robot shifts without requiring manual 3D occupancy annotations.

To this end, we propose a new Hybrid Ray-marching Occupancy Learning Framework called {\ModelName}, and establish {\DatasetName}, a sidewalk perception dataset with LiDAR-camera paired sequences collected across multiple locations and time periods. {\ModelName} predicts both geometry and semantics of sidewalk environments, as shown in Figure~\ref{fig:teaser}. Using a finetuned SAM3~\cite{carion2025sam}, we generate high-quality 3D pseudo-occupancy labels from limited paired camera-LiDAR data to train an initial monocular predictor. We then incorporate large-scale unpaired monocular images through mixed training to increase scene diversity and improve generalization. This combination of pseudo-3D supervision and image-level representation learning provides both geometric grounding and broad visual coverage.
Extensive experiments demonstrate that our approach improves occupancy prediction accuracy and cross-domain generalization. Compared with methods that rely solely on self-supervised image-level signals, our method yields more stable training and greater robustness to both environmental and embodiment shifts. We summarize the main contributions as follows:

    \textbf{1)} We propose \textbf{\ModelName}, a hybrid voxel-ray occupancy perception framework that enables data-efficient learning. By integrating depth-guided ray features with rendering-based self-supervision, {\ModelName} eliminates the need for costly manual 3D annotation and learns to generalize across different camera intrinsics.
    
    \textbf{2)} We introduce \textbf{\DatasetName}, a large-scale, cross-domain RGB-LiDAR dataset specifically for sidewalk robots. {\DatasetName} covers diverse urban environments and time-of-day variations, providing a benchmark for training and evaluating mobile robot perception on real-world, human-centric sidewalks.
    
\textbf{3)} We establish a benchmark on {\DatasetName} to evaluate sidewalk 3D occupancy prediction performance and model generalization across environmental variations and robot cross-embodiment discrepancies. Extensive experiments demonstrate that {\ModelName} achieves state-of-the-art results on this benchmark, with a 15.6\% gain in mIoU. Compared with competitive baselines, our method boosts OOD mIoU by 55\% on the Night split (Set 1), 14\% on the Diverse split (Set 2) , and 3.1\% on cross-embodiment split (Set 3).

\section{Related Work}
We review existing perception datasets for occupancy and sidewalk, as well as advances in semantic scene completion (SSC).
Large-scale LiDAR and multi-view datasets serve as standard benchmarks for occupancy learning and SSC. SemanticKITTI \cite{behley2019semantickitti} provides densely labeled outdoor point clouds, while nuScenes \cite{caesar2020nuscenes} integrates 360° multi-camera and LiDAR data to support multi-view learning. For sidewalk and ground-robot perception, existing datasets have evident limitations: JRDB \cite{martin2021jrdb} focuses on pedestrian detection, RUGD \cite{wigness2019rugd} and Rellis-3D \cite{jiang2021rellis} target off-road semantic segmentation, and long-term SLAM datasets including NCLT \cite{carlevaris2016university} and FusionPortable \cite{jiao2022fusionportable} lack dense semantic labels. SCAND \cite{karnan2022socially} is designed for social navigation without dense perception annotations. In contrast, UT Campus Object Dataset (CODa) \cite{zhang2024toward} provides comprehensive multimodal sidewalk data with 3D bounding boxes and terrain segmentation, well-suited for sidewalk perception tasks. Additionally, recent works including MIMIC \cite{he2026learning} and AURA \cite{ma2026aura} for sidewalk navigation only provide 2D-only datasets.

Early SSC methods depend on labor-intensive dense 3D annotations and computationally costly 3D convolutions. Recent works optimize efficiency via lightweight designs: FlashOcc \cite{yu2023flashocc}, FastOcc \cite{hou2024fastocc}, and SparseOcc \cite{tang2024sparseocc} adopt BEV transformation, cross-feature fusion, and sparse voxel modeling, respectively, to reduce computational overhead. To mitigate annotation costs, label-efficient approaches exploit weak supervision, knowledge distillation, and self-supervision. RenderOcc \cite{pan2024renderocc} and UniOcc \cite{pan2023uniocc} leverage differentiable rendering for 2D-to-3D supervision; RadOcc \cite{zhang2024radocc} applies cross-modal distillation, while OccNeRF \cite{zhang2025occnerf} and SelfOcc \cite{huang2024selfocc} realize LiDAR-free self-supervision using multi-view photometric constraints and pre-trained 2D models. Further moving toward fully unsupervised learning, SceneDINO \cite{jevtic2025feed} lifts self-supervised 2D features to 3D and achieves strong unsupervised SSC performance via multi-view consistency and 3D feature distillation, yet it is not applicable to robots equipped with only monocular cameras.

\section{Framework}
\label{sec:Architecture}

\begin{figure}[t]
    \centering
    \includegraphics[width=\linewidth]{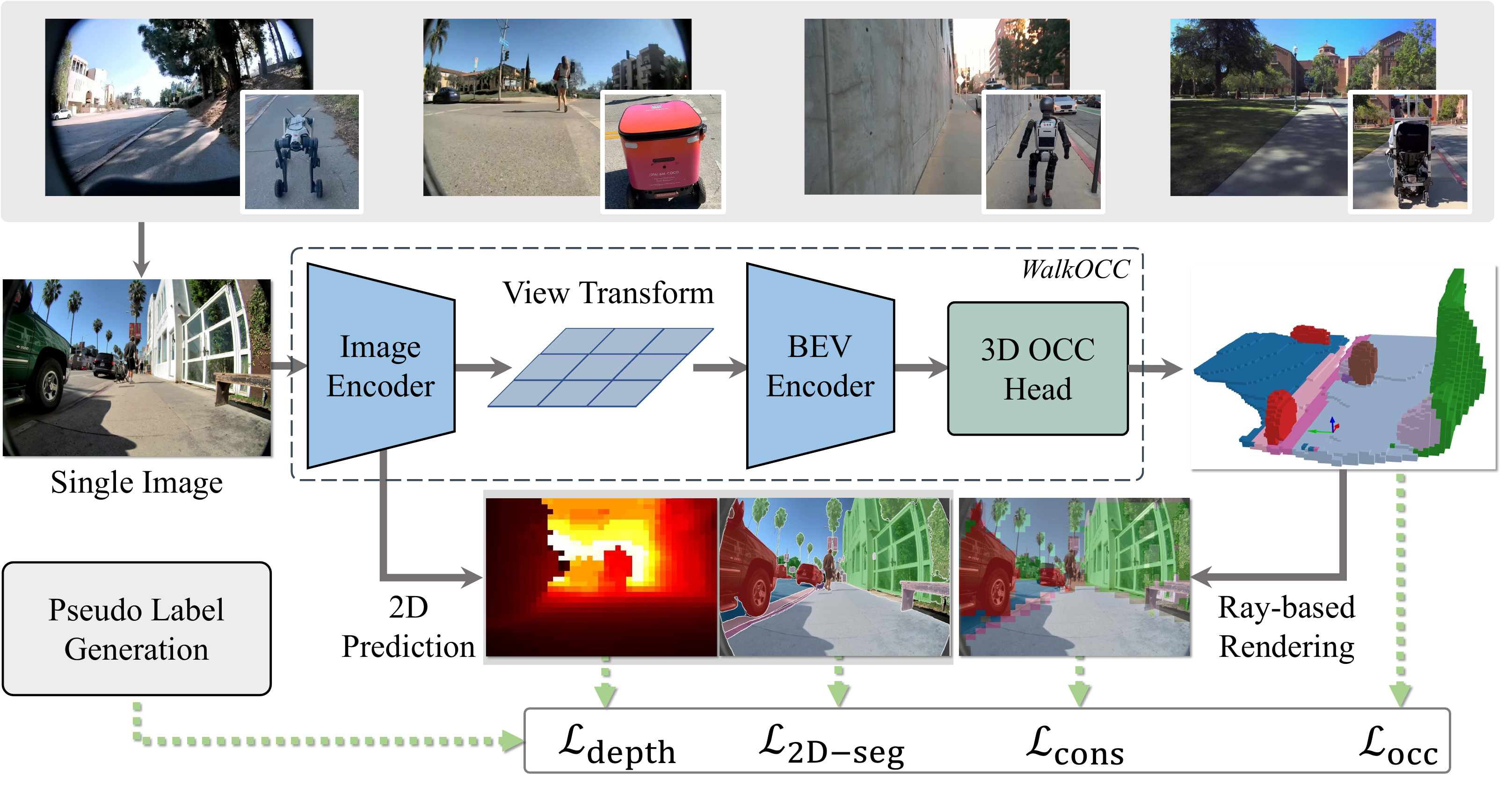}
    \caption{
    \textbf{Learning Framework of {\ModelName}}. {\ModelName} is a lightweight BEV-based occupancy prediction model that takes a single fisheye image as input. The architecture follows an \textbf{Encoder $\rightarrow$ Lift $\rightarrow$ BEV $\rightarrow$ Decoder} paradigm. To facilitate training, we introduce multi-task supervision with auxiliary depth estimation and 2D semantic segmentation. In addition, {\ModelName} adopts a depth-aware lifting module and learns from both 2D and 3D supervision via a ray-marching-based 2D-3D consistency loss. 
    }
    \label{fig:pipeline}
    \vspace{-10pt}
\end{figure}

We present {\ModelName}, a hybrid Ray-marching-based occupancy-learning framework for sidewalk occupancy prediction using a monocular RGB camera. As illustrated in Figure~\ref{fig:pipeline}, our approach consists of two key components: (i) a depth-aware lifting architecture (Section~\ref{sec:LSS}) that transforms front-view images into 3D semantic occupancy grids, and (ii) a hybrid training strategy (Section~\ref{sec:hybrid}) that leverages both 2D and 3D supervision via a ray-marching-based 2D-3D consistency loss. Enforcing this consistency enables effective learning from large-scale 2D-only data while preserving geometric accuracy, which in turn improves prediction quality and cross-domain generalization. After introducing the model, we describe the pseudo-label generation used for training in Appendix~\ref{app:pseudo-label}.

\subsection{Monocular 3D Occupancy via Depth-aware Lifting}
\label{sec:LSS}

Given a monocular RGB image $I_t$ with known camera intrinsics and extrinsics, {\ModelName} predicts a semantic occupancy volume 
$\hat{\mathcal{V}} \in \mathbb{R}^{D \times H_{\mathrm{bev}} \times W_{\mathrm{bev}} \times K}$ 
within a fixed region of interest around the robot. As illustrated in Figure~\ref{fig:pipeline}, the model follows an Encoder--Lift--BEV--Decoder paradigm, which first extracts image-plane features, lifts them into 3D with depth awareness, aggregates them in BEV, and finally decodes dense semantic occupancy.

The image encoder uses a ResNet-50 backbone~\cite{he2016deep} with an FPN-style neck to produce a feature map 
$F \in \mathbb{R}^{C \times H' \times W'}$. 
A lightweight 2D semantic head predicts image-plane logits 
$\hat{S} \in \mathbb{R}^{K \times H \times W}$, 
supervised by ground-truth semantic masks. This auxiliary supervision encourages category-aware image features and also provides the 2D reference for our 2D--3D consistency loss.

To bridge image features and 3D occupancy, we adopt a depth-aware view transformer~\cite{yu2023flashocc,huang2021bevdet,li2023bevdepth}. 
A depth branch predicts a per-pixel depth distribution $\hat{D}(d)$ over discretized depth bins, supervised by LiDAR-projected depth. The 2D features are then lifted along camera rays into a frustum-aligned 3D feature volume:
\begin{equation}
  \mathcal{F}(x,y,z) = \sum_d \hat{D}(d)\,\phi\big(F(u_d, v_d)\big),
\end{equation}
where $(u_d,v_d)$ is the projected image coordinate at depth bin $d$, and $\phi(\cdot)$ denotes a learnable projection. The frustum features are subsequently transformed into a BEV feature map 
$\mathcal{B} \in \mathbb{R}^{C_{\mathrm{bev}} \times H_{\mathrm{bev}} \times W_{\mathrm{bev}}}$ 
and refined by a lightweight BEV encoder.

Finally, a 3D occupancy head decodes the refined BEV representation into semantic voxel predictions. To address the severe imbalance between free space and occupied regions, we use a focal-style voxel-wise occupancy loss:
\begin{equation}
  \mathcal{L}_{\mathrm{occ}}
  =
  \frac{1}{|\Omega|}
  \sum_{v \in \Omega}
  \mathrm{FocalCE}
  \big(
  \hat{\mathcal{V}}_v,
  \mathcal{V}^{\star}_v
  \big),
\end{equation}
where $\Omega$ denotes the valid voxel set and $\mathcal{V}^{\star}_v$ is the ground-truth semantic occupancy label. The full model is jointly supervised by image-plane semantic segmentation, depth estimation, and 3D semantic occupancy losses.

\subsection{2D-3D Consistency Loss and Hybrid Training Strategy}
\label{sec:hybrid}

Existing 2D-3D multi-task methods fail to enforce strict visual alignment for 3D predictions and lack sufficient paired training data, though unpaired 2D semantic data is abundant in robotic environments. To address this gap, we introduce a ray marching-based alignment strategy to align 2D semantic features and 3D occupancy representations along camera rays. This cross-modality constraint regularizes 3D predictions to align with image-based visual cues and enables effective 2D-to-3D knowledge distillation from unpaired 2D data via ray-based semantic propagation.

\begin{wrapfigure}{l}{0.5\columnwidth}
    \vspace{-12pt}
    \includegraphics[width=\linewidth]{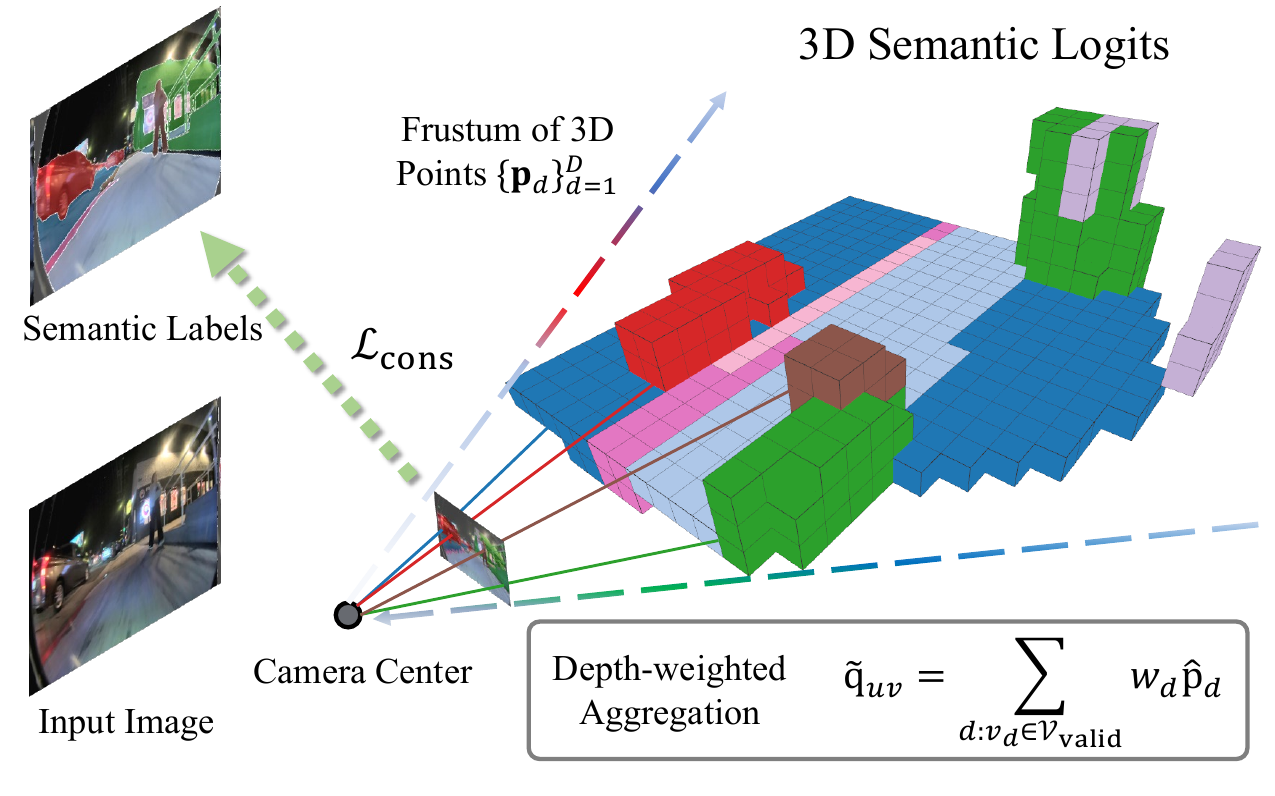}
    \caption{
    \textbf{Ray-based Rendering.}
    }
    \label{fig:rendering}
    \vspace{-10pt}
\end{wrapfigure}
\noindent\textbf{Ray-based 2D-3D semantic alignment via ray marching.}
As illustrated in Figure~\ref{fig:rendering}, ray marching is performed for each image pixel $(u,v)$: starting from the camera center, a viewing ray is cast through the pixel to intersect the 3D occupancy grid. We sample a series of 3D points $\{\mathbf{p}_d\}_{d=1}^{D}$ along the ray, where each point $\mathbf{p}_d$ corresponds to a discrete depth bin $d$. These points are transformed from camera coordinates to ego coordinates via calibrated camera intrinsics and extrinsics, and further mapped to voxel indices $\{v_d\}$ within the occupancy grid.

For each ray-associated voxel $v_d$, we extract the corresponding semantic logits $\hat{\mathbf{p}}_d \in \mathbb{R}^K$ from the predicted 3D occupancy feature $\hat{\mathcal{V}}$. To aggregate these 3D logits into pixel-level 2D semantic predictions, we adopt depth weighting based on the depth distribution $\hat{D}(d)$ predicted by the view transformer, while eliminating interference from background voxels. Let $\mathcal{V}_{\text{valid}}$ denote the set of non-background valid voxels on the ray. The depth-weighted semantic aggregation is formulated as:
\begin{equation}
\tilde{\mathbf{q}}_{uv} = \sum_{d: v_d \in \mathcal{V}_{\text{valid}}} w_d \, \hat{\mathbf{p}}_d,
\end{equation}
where the normalized depth weight $w_d$ masks out background voxels and is defined as:
\begin{equation}
w_d = \frac{\hat{D}(d) \cdot \mathbf{1}[v_d \in \mathcal{V}_{\text{valid}}]}{\sum_{d': v_{d'} \in \mathcal{V}_{\text{valid}}} \hat{D}(d') + \epsilon}.
\end{equation}

The aggregated feature $\tilde{\mathbf{q}}_{uv}$ represents the 3D occupancy-rendered 2D semantic distribution of pixel $(u,v)$, weighted by predicted depth probabilities. We further enforce semantic consistency between this rendered feature and the direct 2D semantic prediction $\hat{\mathbf{q}}_{uv}$ from the semantic head via a consistency loss:
\begin{equation}
\mathcal{L}_{\text{cons}} = \frac{1}{|\Pi|} \sum_{(u,v) \in \Pi} \text{CE}\big(\tilde{\mathbf{q}}_{uv}, \mathbf{q}_{uv}^{\star}\big),
\end{equation}
where $\Pi$ indicates the set of valid non-background pixels, and $\mathbf{q}_{uv}^{\star}$ denotes the pixel-level ground-truth semantic label. This bidirectional consistency constraint tightly couples 2D semantic perception and 3D occupancy estimation, ensuring the model's 3D spatial understanding is strictly aligned with input visual observations.

\paragraph{Hybrid supervision with 2D and 3D labels}
Our training objective combines four complementary terms:
\begin{equation}
  \mathcal{L} = \lambda_{\text{2D}} \mathcal{L}_{\text{2D-seg}} + \lambda_{\text{3D}} \mathcal{L}_{\text{occ}} + \lambda_{\text{cons}} \mathcal{L}_{\text{cons}} + \lambda_{\text{depth}} \mathcal{L}_{\text{depth}}.
\end{equation}
In this hybrid training setup, $\mathcal{L}_{\text{2D-seg}}$ represents a standard cross-entropy or focal loss applied to the 2D semantic masks, while $\mathcal{L}_{\text{occ}}$ supervises the final 3D voxel grid. The depth distributions are regularized by $\mathcal{L}_{\text{depth}}$, and $\mathcal{L}_{\text{cons}}$ constitutes our proposed 2D--3D consistency loss. As a result, each mini-batch conveys a rich mixture of complementary signals: dense 2D supervision at the image level, sparse but highly informative 3D occupancy labels, ray-wise consistency constraints that bind them together, and geometric depth supervision that stabilizes the lifting process.

\section{Experiments}
\label{sec:exp}
We construct a benchmark for sidewalk occupancy prediction in Section~\ref{exp:benchmark} and evaluate the out-of-domain (OOD) generalization performance in Section~\ref{exp:ood}. Additionally, we analyze the impact of occupancy supervision on trajectory prediction in Appendix~\ref{app:downstream}. Please refer to Appendix~\ref{app:ablation} for additional ablation studies.

\subsection{Sidewalk Occupancy Prediction Benchmark}
\label{exp:benchmark}

\begin{table*}[t]
\small
\centering
\setlength{\tabcolsep}{0.0025\linewidth}
\renewcommand{\arraystretch}{1.0}

\caption{Comparison of 3D occupancy prediction methods. Bold entries indicate the best performance.}
\label{table:base_main}
\resizebox{\linewidth}{!}{
\begin{tabular}{l c c c c c c c c c c c c c c c c}
\toprule
Method
& \makecell[c]{Input}
& \makecell[c]{occ\_IoU}
& \makecell[c]{mIoU}
& \rotatebox{90}{\textcolor{road}{$\blacksquare$} road} 
& \rotatebox{90}{\textcolor{sidewalk}{$\blacksquare$} sidewalk}
& \rotatebox{90}{\textcolor{crosswalk}{$\blacksquare$} crosswalk} 
& \rotatebox{90}{\textcolor{grass}{$\blacksquare$} grass} 
& \rotatebox{90}{\textcolor{wall}{$\blacksquare$} wall} 
& \rotatebox{90}{\textcolor{trees}{$\blacksquare$} trees} 
& \rotatebox{90}{\textcolor{vehicle}{$\blacksquare$} vehicle} 
& \rotatebox{90}{\textcolor{cyclist}{$\blacksquare$} cyclist} 
& \rotatebox{90}{\textcolor{animal}{$\blacksquare$} animal} 
& \rotatebox{90}{\textcolor{generic_obstacle}{$\blacksquare$} obstacle} 
& \rotatebox{90}{\textcolor{pedestrian}{$\blacksquare$} pedestrian} 
& \rotatebox{90}{\textcolor{curb}{$\blacksquare$} curb} 
& \rotatebox{90}{\textcolor{gutter}{$\blacksquare$} gutter} \\
\midrule
Pseudo Label & C+L   & 88.98 & 56.71
& 69.42 & 80.73 & 61.39 & 72.98 & 79.59 & 65.83 & 79.78 & 13.68 & 43.38 & 26.07 & 64.37 & 42.74 & 37.3 \\
\midrule
{GaussianOcc}~\cite{gan2025gaussianocc}  & C  & 16.29 & 5.60
& 8.72 & 21.86 & 9.16 & 9.52 & 5.65 & 3.99 & 3.13 & 0.06 & 0.0 & 0.87 & 0.53 & 4.21 & 5.04 \\
{MonoScene}~\cite{cao2022monoscene} & C  & 23.08 & 8.42
& 10.30 & 39.80 & 19.64 & 9.04 & 6.32 & 4.06 & 1.96 & 0.16 & 0.0 & 1.40 & 0.74  & 3.46 & 4.12 \\
{TPVFormer}~\cite{huang2023tri} & C & 22.89 & 9.39
& 14.10 & 30.14 & 17.64 & 15.30 & 7.25 & 6.10 & 6.60 & 0.44 & 0.10 & 2.89 & 3.00 & 7.20 & 11.34 \\
{RenderOCC}~\cite{pan2024renderocc} &  C & 25.58 & 13.03
& 19.35 & 34.05 & 22.71 & 19.84 & 15.43 & 9.72 & 11.25 & 2.65 & 0.12 & 6.00 & 7.27 & 9.61 & 11.40 \\
{FlashOCC}~\cite{yu2023flashocc} & C & 27.17 & 14.23
& 21.12 & 35.46 & 24.37 & 21.58 & 16.38 &\textbf{ 10.24 }& 14.12 & 3.27 & 0.14 & 6.76 & 8.82 & {10.64} & 12.04 \\
\midrule
{{\ModelName}} & C & \textbf{30.0}2 & \textbf{16.46}
& \textbf{23.55} & \textbf{38.11} & \textbf{27.61} & \textbf{25.40} & \textbf{17.41} & 9.84 & \textbf{18.74} & \textbf{4.15} & \textbf{0.21} & \textbf{7.53} & \textbf{14.59} & \textbf{12.87} & \textbf{13.96} \\
\bottomrule
\end{tabular}
}
\end{table*}
\paragraph{Implementation details}
\begin{figure}[t]
    \centering
    \includegraphics[width=\linewidth]{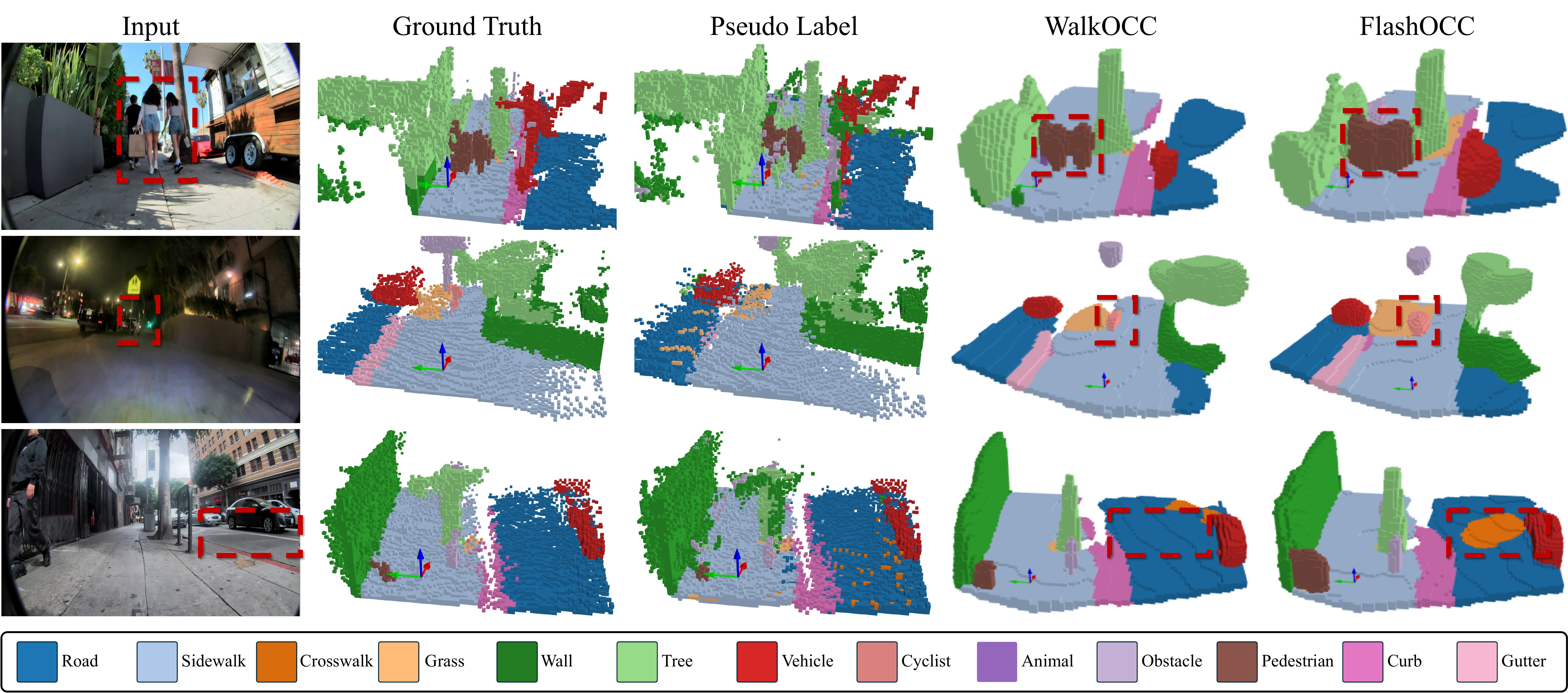}
    \caption{
    \textbf{Qualitative results on the {\DatasetName} dataset}. We present three inference results for our method and FlashOCC on the {\DatasetName} test set, along with the ground truth and pseudo labels for reference. Our predictions are more accurate and exhibit clearer boundaries.
    }
    \label{fig:vis}
\end{figure}
We conduct all experiments over 24 epochs using 8 NVIDIA A5000 GPUs. 3D semantic occupancy prediction is evaluated via mIoU on 14 classes. We compute mIoU based on foreground classes only and additionally report binary occ\_IoU for occupied versus free space. The ego-frame occupancy grid has a $0.1$~m voxel size and forms a $100 \times 100 \times 50$ grid over $X \in [0,10]$~m, $Y \in [-5,5]$~m and $Z \in [-2,3]$~m. Depth is discretized within $[0.2,12]$~m with a step size of $0.05$~m. Front-facing camera inputs are resized from $1080 \times 1920$ to $544 \times 960$.

\paragraph{Baselines}

We compare with five representative monocular occupancy baselines that cover complementary design goals for sidewalk robots: MonoScene~\cite{cao2022monoscene} as a strong and widely used monocular semantic scene completion method, RenderOcc~\cite{pan2024renderocc} as a label-efficient approach that unifies 2D/3D supervision via rendering-based training, FlashOcc~\cite{yu2023flashocc} as a lightweight design optimized for fast, low-memory onboard inference, GaussianOCC~\cite{gan2025gaussianocc} leveraging Gaussian representation for high-fidelity occupancy prediction, and TPVFormer~\cite{huang2023tri} adopting tri-plane vision transformer for robust spatial feature modeling.

\paragraph{Quantitative Results}
Before comparing methods, we first evaluate the quality of the pseudo-labels used to build our benchmark (first row of Table~\ref{table:base_main}). The pseudo-labels closely match a manually verified subset, suggesting that they are sufficiently accurate to provide reliable supervision in this sidewalk-robotics setting. This supports using our pseudo-label-driven pipeline as a reasonable baseline for this benchmark. Table~\ref{table:base_main} presents the main results on our sidewalk occupancy benchmark across five monocular baselines. Our {\ModelName} achieves the best overall performance, improving occ\_IoU from 27.17 to 30.02 and mIoU from 14.23 to 16.46 compared with the strongest baseline FlashOcc~\cite{yu2023flashocc}. The gains are consistent across most semantic categories, with particularly notable improvements on dynamic and safety-critical classes such as vehicle (14.12,$\rightarrow$,18.74), cyclist (3.27,$\rightarrow$,4.15), and pedestrian (8.82,$\rightarrow$,14.59). Our method also improves performance on common traversable regions, including road (21.12,$\rightarrow$,23.55), sidewalk (35.46,$\rightarrow$,38.11), and crosswalk (24.37,$\rightarrow$,27.61), as well as on grass, wall, obstacle, curb, and gutter. The only exception is trees, where FlashOcc remains slightly better (10.24 vs.\ 9.84), suggesting that thin vegetation structures remain challenging under monocular supervision.

\paragraph{Qualitative Results}
As shown in Figure~\ref{fig:vis}, we present qualitative comparisons among the ground truth, pseudo labels, our {\ModelName}, and FlashOcc (from left to right). We visualize four test samples: two touristy scenes during daytime (first row), the same location at night (second row), and a commercial district scene (third row). Overall, our {\ModelName} yields predictions that better match the ground truth, while substantially reducing the noise and artifacts observed in the pseudo labels and the baseline results.

\subsection{Cross-Domain Generalization Evaluation}
\label{exp:ood}
Sidewalk scenes exhibit substantial appearance and structural variations across {locations, capture conditions (e.g., daytime vs.\ nighttime), and robot embodiments with different camera configurations}. To evaluate OOD generalization, we train all models on a single source domain (touristy, daytime; 2.4K samples) and directly test them on target domains without using any target data or adaptation. 
{We define three distinct OOD evaluation settings. The first two correspond to environmental distribution shifts: OOD Set 1 comprises touristy scenes captured under nighttime conditions, while OOD Set 2 includes all remaining test samples apart from the touristy-daytime partition. The third setting targets cross-embodiment distribution shifts, in which OOD Set 3 is collected from diverse platforms, including humanoid, quadruped, and wheeled robots with varying camera intrinsics and mounting angles. Since this dataset lacks paired 3D LiDAR annotations, we convert predicted 3D occupancy volumes into 2D representations via spatial projection and adopt pixel accuracy and 2D mIoU as the evaluation metrics for quantitative assessment.}

\begin{figure}[t]
  \centering
  \begin{subfigure}{0.48\linewidth}
    \centering
    \includegraphics[width=\linewidth]{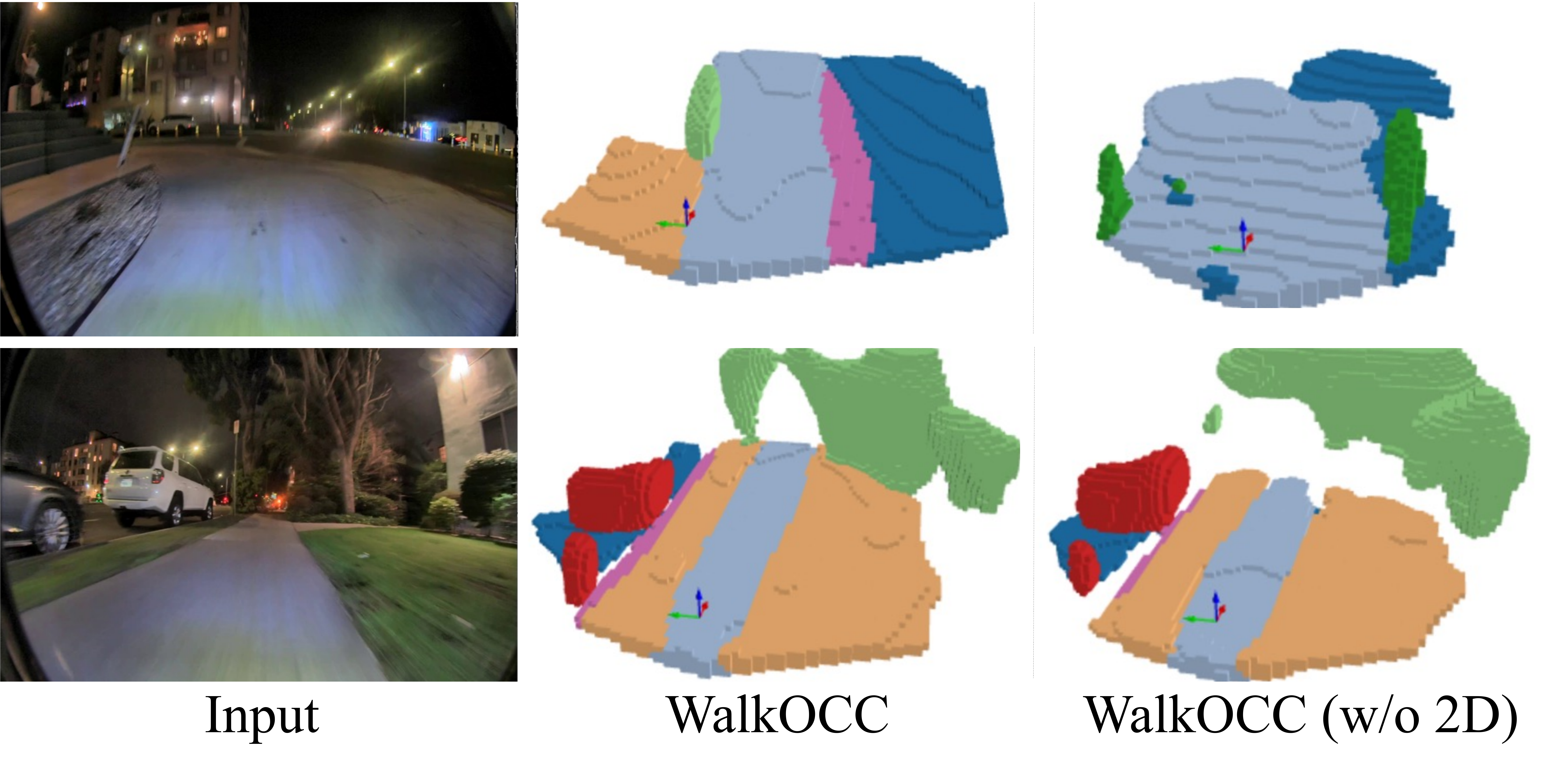}
    \caption{Environmental shift} 
  \end{subfigure}
  \hfill
  \begin{subfigure}{0.48\linewidth}
    \centering
    \includegraphics[width=\linewidth]{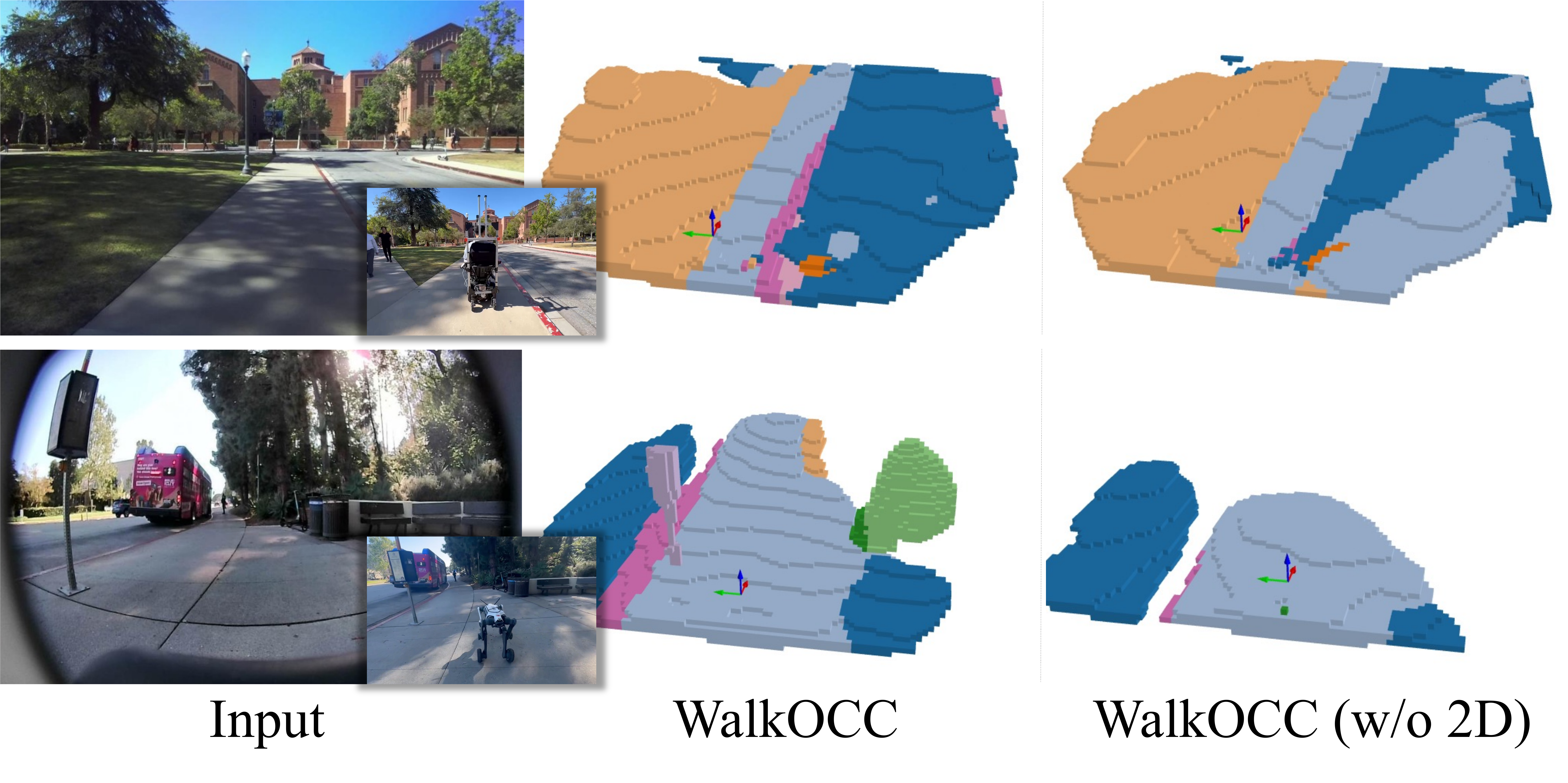}
    \caption{Embodiment shift} 
  \end{subfigure}
  \caption{\textbf{OOD qualitative comparison.} Hybrid training with 2D extended data produces more distinct road structures and more accurate object recognition under both environmental and embodiment shifts.}
  \label{fig:OOD}
  \vspace{-15pt}
\end{figure}

Table~\ref{tab:cross_domain} summarizes the results using mIoU and occupancy IoU (Occ IoU); we additionally report relative mIoU (Rel-mIoU), i.e., the percentage of source-domain mIoU retained on the target domain. Our {\ModelName} incorporates extra 2D-only supervision (2.2K samples) from MIMIC~\cite{he2026learning} for hybrid training, which noticeably improves robustness under domain shift. Concretely, {\ModelName} improves OOD mIoU from 5.55 to 8.61 on set~1 and from 9.03 to 10.29 on set~2 compared to FlashOCC, while also achieving higher Occ IoU (13.77$\rightarrow$18.03 on set~1 and 18.8$\rightarrow$20.11 on set~2). Notably, adding the 2D-only data yields a large gain over {\ModelName} w/o 2D on the more challenging nighttime split (mIoU 5.70$\rightarrow$8.61), indicating that the proposed hybrid training helps preserve semantic discrimination when visual conditions change. Overall, {\ModelName} retains 45.6\%/54.5\% of its in-domain mIoU on OOD set~1/set~2, outperforming the baselines and suggesting better generalization to unseen domains.

Furthermore, we provide qualitative visualizations to compare models trained with and without hybrid training, as shown in Fig.~\ref{fig:OOD}(a). We report OOD examples, comparing our model trained with and without the 2D extension data. The model without hybrid training fails in night scenes: it can no longer delineate the road surface or recognize distant obstacles. In contrast, the hybrid-trained model preserves clearer road structure and can still identify far-away trees. This setting is particularly challenging for urban perception because it contains no night-time training data.

\begin{wraptable}{r}{0.48\linewidth}
  \centering
  \vspace{-10pt}
  \caption{3D occupancy projected to 2D semantic evaluation on robot test set.
  Metrics are computed only on pixels whose GT depth ray endpoint lies inside the OCC grid (same mask as $\mathcal{L}_{\mathrm{cons}}$).}
  \label{tab:ood3}
  \resizebox{\linewidth}{!}{
  \begin{tabular}{lcc}
    \toprule
    Method & Pixel Acc (\%) $\uparrow$ & mIoU (\%) $\uparrow$ \\
    \midrule
    {\ModelName} w/o 2D  & 71.90 & 24.73 \\
    {\ModelName} & \textbf{73.8}    & \textbf{25.5 }   \\
    \bottomrule
  \end{tabular}
  }
\vspace{-10pt}
\end{wraptable}

As shown in Tab.~\ref{tab:ood3}, we evaluate the model generalization to cross-embodiment shifts with distinct camera intrinsic parameters (OOD Set 3). As this evaluation set lacks 3D LiDAR annotations, we project all predicted occupancy volumes onto the 2D image plane and report pixel accuracy and 2D mIoU instead of 3D mIoU. {\ModelName} trained with cross-embodiment 2D images consistently surpasses the baseline variant trained without such data across both metrics, demonstrating that our 2D-3D hybrid training paradigm effectively addresses intrinsic camera variations without any extra 3D supervision, as shown in Fig.~\ref{fig:OOD}(b). For more details and visualizations, please refer to the Appendix~\ref{app:vis}.



\begin{table}[t]
\centering
\caption{Cross-domain evaluation of SSC under different test conditions. Models are trained on the source domain and evaluated on three target conditions. Rel-mIoU denotes the relative mIoU compared to source-domain performance. Occ\_IoU measures occupancy completion by collapsing semantic labels into occupied and free space. }
\label{tab:cross_domain}
\resizebox{\linewidth}{!}{
\begin{tabular}{c|cc|ccc|ccc}
\toprule
\multirow{2}{*}{Method}
& \multicolumn{2}{c|}{In Domain}
& \multicolumn{3}{c|}{Out of Domain set 1}
& \multicolumn{3}{c}{Out of Domain set 2} \\
\cline{2-9}
& mIoU $\uparrow$
& Occ\_IoU $\uparrow$
& mIoU  $\uparrow$
& Rel-mIoU $\uparrow$
& Occ\_IoU $\uparrow$
& mIoU  $\uparrow$
& Rel-mIoU $\uparrow$
& Occ\_IoU $\uparrow$ \\
\midrule
FlashOCC
& 17.13 & 29.18
& 5.55  &  32.39 &  13.77
& 9.03 & 52.71\%  & 18.8 \\
{{\ModelName} w/o 2D}
&  18.78 & \textbf{31.08}
& 5.70  & 30.35\% & 11.97
& 9.25 & 49.25\% & 17.77 \\
{\ModelName}
& \textbf{18.88} & {30.79}
& \textbf{8.61} & \textbf{45.6\%} & \textbf{18.03} 
& \textbf{10.29} & \textbf{54.50\%} & \textbf{20.11} \\
\bottomrule
\end{tabular}
}
\vspace{-10pt}
\end{table}

\section{Conclusion}
We present {\ModelName}, a hybrid ray-marching framework for monocular sidewalk occupancy that bootstraps pseudo 3D supervision from limited RGB--LiDAR pairs and scales with unpaired 2D images via depth-aware lifting and 2D--3D consistency.
Together with the {\DatasetName} benchmark, experiments show improved in-domain accuracy, stronger robustness to environmental and cross-embodiment shift, and finer structure recovery than competitive monocular baselines.

\section{Limitations}
{\ModelName} assumes accurate camera calibration, a fixed occupancy volume, and reliable pseudo 3D labels from limited paired RGB--LiDAR data; errors or strong domain shift (e.g., nighttime or unseen embodiments) can blur boundaries and hurt rare or distant objects.
Future work will expand data and onboard computing efficiency.

\clearpage


\bibliography{example}  

\clearpage

\appendix

\begin{center}
    \LARGE \textbf{Appendix}
\end{center}
\begin{abstract}
This supplementary material presents additional evidence and comprehensive implementation details. It covers the {effect of occupancy supervision on trajectory planning (Section~\ref{app:downstream})}, further ablation studies (Section~\ref{app:ablation}), qualitative and cross-embodiment visualizations (Section~\ref{app:vis}), a complete description of the pseudo-label generation pipeline (Section~\ref{app:pseudo-label}), fine-tuning details for the sidewalk-specialized SAM3 (Section~\ref{app:sam3}), and dataset collection and preprocessing procedures (Section~\ref{app:data-preprocess}).
\end{abstract}

{
\startcontents[sections]
\noindent\textbf{Contents}\par\smallskip
\printcontents[sections]{l}{1}{\setcounter{tocdepth}{3}}
\medskip
}

{

\section{Occupancy-aware Trajectory Planning}\label{app:downstream}
As discussed in the introduction, dense occupancy provides a geometric prior well suited to legged navigation: predicted future motion can be checked against walkable surfaces, obstacles, and terrain classes rather than sparse 2D cues alone.
We attach a lightweight trajectory decoder to the same BEV feature that feeds the occupancy head, supervise it with future ego poses, and evaluate paths with occupancy-grounded metrics on the COCOa test split.

\subsection{Trajectory Planning Head}
\label{app:traj_head}

We supervise ten future ego-frame waypoints $(x,y)$ at 2\,Hz (5\,s horizon), with $x$ forward and $y$ left; frames without a full future window are excluded.
During training, these waypoints undergo the same BEV rotation, scaling, and flipping as the occupancy grid.

The trajectory head reads the shared BEV feature $\mathbf{F} \in \mathbb{R}^{B \times C \times D_y \times D_x}$ ($C{=}128$) used by the occupancy decoder: two $3{\times}3$ convolutions (stride~2 on the second), $4{\times}4$ adaptive average pooling, and a three-layer MLP output $\hat{\mathbf{P}} \in \mathbb{R}^{B \times T \times 2}$.
Training uses Smooth-L1 on valid 3D samples.
By default, occupancy, depth, semantic, and trajectory losses all update the encoder.
For \textbf{Ours w/o OCC supervision} in Table~\ref{tab:traj_nav_occ}, only the trajectory loss back-propagates; \textbf{Ours w/ OCC supervision} restores joint training for the navigation metrics below.

\subsection{Metrics and Results}
\label{sec:traj_nav_metrics}

We evaluate predicted future trajectories with three navigation-aware rates derived from dense GT occupancy in the ego BEV. The 3D voxel labels are collapsed to a 2D semantic map: for each $(x,y)$ column we take a majority vote over all non-background voxels whose vertical center satisfies $z \le z_{\mathrm{ceil}}$ (we use $z_{\mathrm{ceil}}{=}1.0$\,m).
Each waypoint $(x_t,y_t)$ is tested with a circular ego footprint of radius $r{=}0.4$\,m in the BEV plane.
Let $\mathcal{C}$ denote obstacle classes (\texttt{wall}, \texttt{trees}, \texttt{vehicle}, \texttt{cyclist}, \texttt{animal}, \texttt{obstacle}, \texttt{pedestrian}).
A timestep counts as a \textbf{collision} if any cell under the footprint belongs to $\mathcal{C}$; if the disk is entirely background, we use the nearest non-background BEV cell as the ruling label.
A timestep counts as \textbf{grass contact} if any footprint cell is \texttt{grass}.
\textbf{Lane keeping} is defined for sidewalk-centric locomotion: a timestep is lane-ok iff there is no collision and the ruling label is not \texttt{road} (motor-vehicle surface); sidewalk, crosswalk, curb, and similar non-road classes do not incur a violation.
Per trajectory we report the fraction of timesteps in each category; dataset-level \textbf{collision rate}, \textbf{grass rate}, and \textbf{lane-keeping rate} are the mean of these fractions over all test samples with a valid prediction.

Table~\ref{tab:traj_nav_occ} reports the three rates on our test set.
Without occupancy supervision (\emph{w/o} OCC), we obtain 2.79\% collision, 0.47\% grass contact, and 95.03\% lane keeping.
With dense OCC supervision (\emph{w/} OCC), all metrics improve: collision \textbf{2.37\%} ($\downarrow$0.42\,pp), grass \textbf{0.31\%} ($\downarrow$0.16\,pp), and lane keeping \textbf{95.43\%} ($\uparrow$0.40\,pp).
Because scoring uses GT semantics as a fixed oracle, these numbers measure whether predicted paths respect walkable structure (obstacle avoidance, staying off grass, and avoiding motor-road surfaces) independently of voxel prediction accuracy reported in other tables.
\begin{table}[t]
  \centering
  \caption{Trajectory metrics against dense occupancy supervision in the ego BEV (GT OCC, $r{=}0.4$\,m footprint, $z_{\mathrm{ceil}}{=}1.0$\,m).}
  \label{tab:traj_nav_occ}
  \begin{tabular}{lccc}
    \toprule
    \textbf{Method} &
    Collision rate (\%) $\downarrow$ &
    Grass rate (\%) $\downarrow$ &
    Lane-keeping rate (\%) $\uparrow$ \\
    \midrule
    Ours \emph{w/o} OCC supervision & 2.79 & 0.47 & 95.03 \\
    Ours \emph{w/} OCC supervision & \textbf{2.37} & \textbf{0.31} & \textbf{95.43} \\
    \bottomrule
  \end{tabular}
\end{table}

\begin{figure}[H]
    \centering
    \includegraphics[width=0.95\linewidth]{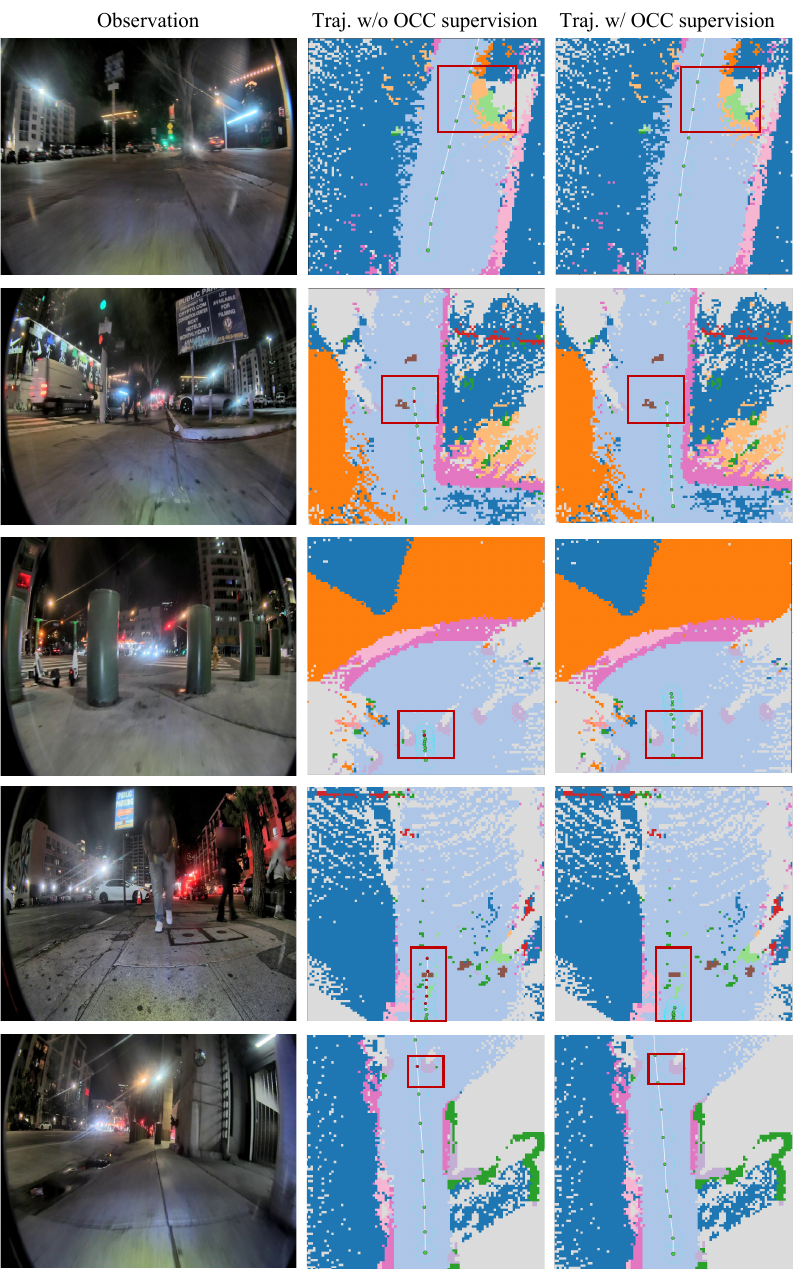}
    \caption{
    Comparison of trajectory prediction results with and without occupancy supervision. From left to right, we present the current observation, trajectory prediction without OCC supervision, and trajectory prediction with OCC supervision. Predicted trajectories are visualized as green waypoints connected by white lines. Red waypoints indicate collisions between the occupancy area of the ego vehicle (marked as a blue circle) and surrounding obstacles.
    }
    \label{fig:addcollision}
\end{figure}

\subsection{Visualization}
As visualized in Figure~\ref{fig:addcollision}, we overlay predicted trajectories onto the occupancy ground truth from a top-down perspective. The results reveal that occupancy supervision provides reliable prior information for trajectory forecasting, effectively reducing occurrences of driving onto grassland and collisions with pedestrians and obstacles. Such advantages are particularly prominent in densely crowded scenarios, as illustrated in the third and fourth rows. Meanwhile, occupancy cues also improve trajectory prediction performance for distant regions, shown in the first and fifth rows.

}

\section{Additional Ablation Studies}\label{app:ablation}

\subsection{Ablation of Model Components.}
\label{exp:abl}
Table~\ref{tab:ablation} reports an ablation of the training objectives and the proposed hybrid mixed supervision (MS). All variants are trained on the {\DatasetName} \emph{train} split and evaluated on the {\DatasetName} \emph{test} split.

\begin{wraptable}{r}{0.48\linewidth}
\vspace{-15pt}
\centering
\caption{Ablation study of the proposed loss terms and mixed supervision (MS). Checkmarks indicate the corresponding component is enabled.}
\label{tab:ablation}
\resizebox{\linewidth}{!}{%
\begin{tabular}{cccc|cc}
\toprule
$\mathcal{L}_\text{depth}$ &$\mathcal{L}_\text{2D-seg}$ & $\mathcal{L}_\text{cons}$ & MS & IoU & mIoU  \\
\midrule
  &        &        &  &  14.09 & 24.54 \\
\checkmark &  &  &  & 14.23 & 27.17   \\
\checkmark & \checkmark &  &  & 15.71  & 29.09   \\
\checkmark & \checkmark & \checkmark &  & 16.32 & 29.87  \\
\checkmark & \checkmark & \checkmark & \checkmark & \textbf{16.46} & \textbf{30.02}  \\
\bottomrule
\end{tabular}%
}
\vspace{-10pt}
\end{wraptable}

Starting from the baseline model without auxiliary objectives, adding depth supervision ($\mathcal{L}_\text{depth}$) yields a clear improvement in both Occ IoU and mIoU, indicating better geometric completion. Introducing 2D semantic supervision ($\mathcal{L}_\text{2D-seg}$; semantic loss, SL) further boosts the scores by enhancing semantic discrimination. We then enable the 2D-3D consistency loss ($\mathcal{L}_\text{cons}$; consistency loss, CL), which provides an additional gain, suggesting that aligning predictions across modalities helps reduce ambiguous label assignments.
Finally, we activate hybrid training with extra 2D-only data (MS). The improvement is marginal on the in-domain setting, which is expected: the full-size {\DatasetName} training set already largely covers the test-domain distribution, so mixing in more diverse 2D data brings limited additional benefit here.

\subsection{Ablation on Occupancy Grid Resolution}
We train and evaluate the model on the touristy daytime subset.
We evaluate the impact of voxel resolution and spatial range using multiple complementary metrics.
We report the standard mIoU to measure overall semantic scene completion performance.
To decouple geometric completion from semantic prediction, we additionally report occupancy IoU (Occ IoU), where all semantic classes are merged into occupied and free space.
The results in Table~\ref{tab:range_resolution} show a clear resolution--range trade-off. Under a fixed spatial range of 10$\times$10$\times$5~m, increasing the voxel size from 0.05~m to 0.20~m consistently improves both semantic completion (mIoU: 10.44 $\rightarrow$ 20.52) and geometry completion (Occ IoU: 20.24 $\rightarrow$ 41.18). We attribute this to the fact that coarser grids reduce the difficulty of predicting fine-grained structure and are less sensitive to sensor noise and label ambiguity, thus yielding higher overlap scores under our evaluation protocol. In contrast, enlarging the spatial range from 10$\times$10$\times$5~m to 20$\times$20$\times$5~m at the same 0.10~m resolution leads to a noticeable drop (mIoU: 16.25 $\rightarrow$ 11.35; Occ IoU: 29.09 $\rightarrow$ 21.84), likely because distant regions are sparser and more occluded, which increases the fraction of hard-to-complete voxels and dilutes supervision.

\begin{table}[t]
\centering
\begin{minipage}{0.48\linewidth}
\centering
\caption{Impact of voxel resolution and spatial range on SSC performance.}
\label{tab:range_resolution}
\begin{tabular}{cc|cc}
\toprule
Range (m) & Res. (m) & mIoU $\uparrow$ & OCC IoU $\uparrow$   \\
\midrule
10$\times$10$\times$5 & 0.05 & 10..44 & 20.24 \\
10$\times$10$\times$5 & 0.10 &  16.25& 29.09 \\
10$\times$10$\times$5 & 0.20 & 20.52 & 41.18 \\
\midrule
20$\times$20$\times$5 & 0.10 & 11.35 & 21.84 \\
\bottomrule
\end{tabular}
\end{minipage}
\hfill
\begin{minipage}{0.48\linewidth}
\centering
\caption{Impact of OOD data construction.}
\label{tab:percent}
\begin{tabular}{c|cc}
\toprule
Nighttime (\%) & mIoU $\uparrow$ & OCC IoU $\uparrow$ \\
\midrule
w/o 2D & 6.13& 15.82\\
\midrule
30 & 8.74& 19.72\\
60 & 9.93& 20.48\\
90 & 8.66& 18.33\\
\bottomrule
\end{tabular}
\end{minipage}
\end{table}

\subsection{Ablation on Target Data}
We train the model on the touristy daytime subset but test it on the touristy nighttime out-of-domain subset. We keep the 2D extension data fixed to 5K samples, and vary the proportion of nighttime target data to 30\%, 60\%, and 90\% to study how the target-data composition affects training. As shown in Table~\ref{tab:percent}, increasing the nighttime ratio from 30\% to 60\% improves the results, while further increasing it to 90\% leads to a consistent drop. We attribute this degradation mainly to a \,\emph{bias--diversity trade-off}: with 90\% nighttime scenes, the model is exposed to much less appearance diversity (e.g., fewer well-lit daytime examples), making the learned features overfit to the specific nighttime photometric statistics and thus less robust at test time. In addition, nighttime data typically contains noisier visual cues (low illumination, motion blur, headlight flare) and sparser/less reliable pseudo labels, which can amplify label noise during training when it dominates the batches. Finally, the fixed 2D extension set may have a different class/context distribution compared with heavily night-biased target data, creating supervision conflicts that hinder stable optimization and hurt overall semantic completion performance.

\section{Additional Visualizations}\label{app:vis}
In this section, we provide two additional visualizations. First, in Section~\ref{sec:addvismodel}, we present more qualitative inference results of {\ModelName} on {\DatasetName} to highlight the model's performance in challenging scenarios and the diversity of our dataset. Second, in Section~\ref{sec:cross}, we evaluate the model in a zero-shot manner across different embodiments to demonstrate its generalizability and the practical value of deploying micro-mobility systems on sidewalks.
\subsection{Additional Inference Visualizations with {\ModelName}}
\label{sec:addvismodel}

\begin{figure}[H]
    \centering
    \includegraphics[width=\linewidth]{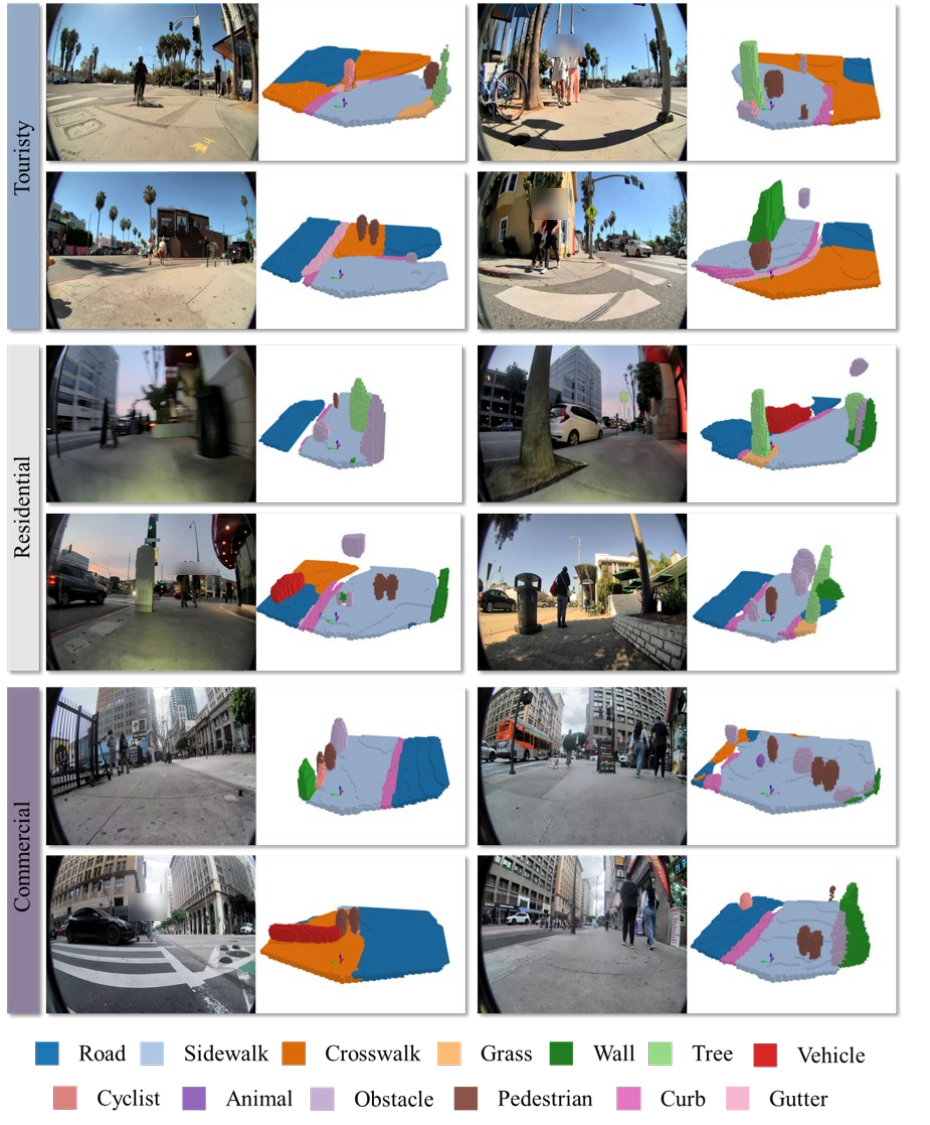}
    \caption{
        \textbf{Qualitative inference results of {\ModelName} across scene types.} For each subset of {\DatasetName} (touristy, residential, and commercial), we show four examples with the input image and the predicted occupancy. {\ModelName} consistently recovers the static scene layout and localizes dynamic agents, even in crowded touristy streets and wide commercial roads.
    }
    \label{fig:Additionalvis}
\end{figure}

As shown in Figure~\ref{fig:Additionalvis}, we present four examples from each subset of our dataset: touristy, residential, and commercial. Touristy scenes are typically crowded with pedestrians, highlighting the model's ability to predict pedestrian motion under dense interactions. Commercial scenes often feature heavier traffic and wider sidewalks/roads, demonstrating that the model can also capture the structural layout of the environment. Residential scenes lie between these two extremes, suggesting that the model generalizes well across different appearances and spatial layouts.

\subsection{Cross-Embodiment Inference}
\label{sec:cross}

\begin{figure}[H]
    \centering
    \includegraphics[width=\linewidth]{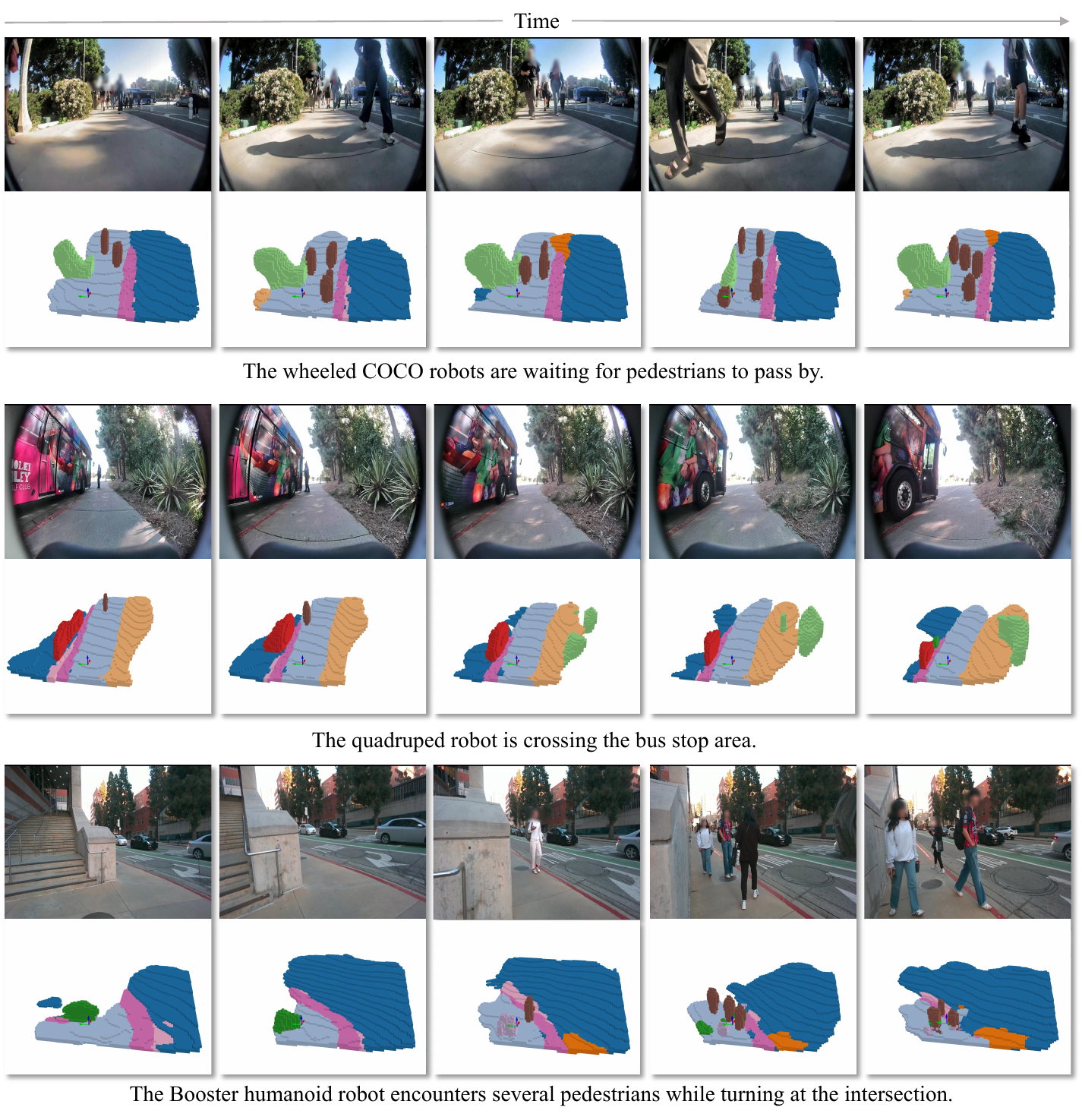}
    \caption{\textbf{Qualitative cross-embodiment results.} We zero-shot deploy our perception model on three representative platforms (wheeled, quadruped, and humanoid) with different camera heights and intrinsics. Across typical sidewalk scenarios (yielding to pedestrians, traversing a bus-stop area, and turning at an intersection), the model exhibits strong open-world cross-embodiment generalization, indicating its promise for real-world sidewalk deployment.}
    \label{fig:vis_video}
\end{figure}

Different from autonomous driving, sidewalks involve diverse micro-mobility devices. Here, we perform zero-shot inference with our perception model to evaluate its cross-embodiment generalizability. As shown in Figure~\ref{fig:vis_video}, we present three representative platforms: a wheeled robot, a quadruped robot, and a humanoid robot. These examples include typical scenarios such as stopping and waiting for pedestrians to pass, traversing a bus-stop area, and turning at an intersection. The results demonstrate strong cross-embodiment generalization in open-world settings, highlighting the practical value of our approach for real-world deployment.

\section{Pseudo-Label Pipeline Details}
\label{app:pseudo-label}

\begin{figure}[t]
    \centering
    \includegraphics[width=\linewidth]{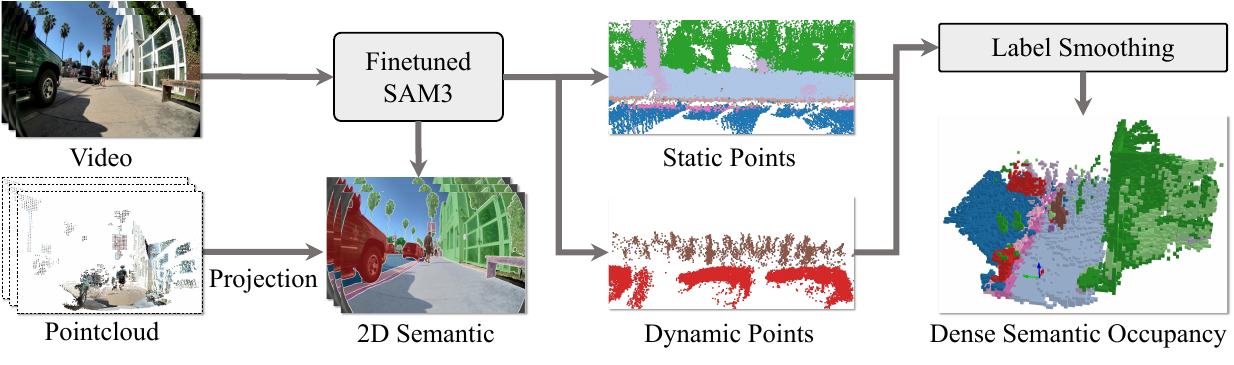}
    \caption{
    \textbf{Pseudo-Label Generation}.
    With pre-calibrated and time-synchronized sensors, we project 3D LiDAR points onto 2D images to inherit per-point semantic labels. We then generate dense occupancy pseudo-labels using the SurroundOcc~\cite{wei2023surroundocc} pipeline, taking as input the static-scene point cloud and dynamic objects. To improve label quality in sidewalk-centric scenes, we finetune SAM3~\cite{carion2025sam} and apply label smoothing to mitigate noisy predictions.
    }
    \label{fig:pseudo}
\end{figure}

This section provides implementation details for Section~3 of the main paper.
Figure~\ref{fig:pseudo} summarizes the pipeline, and Algorithm~\ref{fig:pseudo-pipeline-pseudocode} formalizes the procedure.
Given a synchronized LiDAR--camera sequence, we: (i) run SAM3 with an overcomplete prompt set to obtain dense 2D masks (Section~\ref{sec:overcomplete-prompts}); (ii) merge base and fine-tuned passes into a single per-pixel label map (Section~\ref{sec:overcomplete-prompts}); (iii) lift labels to 3D by projecting LiDAR points into the front-view image (Section~\ref{sec:3d-lift}); (iv) optionally enforce temporal/spatial consistency via 3D voting (Section~\ref{sec:voting-flicker}); and (v) convert the labeled point clouds to the SurroundOcc format to generate semantic occupancy ground truth (Section~\ref{sec:surroundocc-gt}).

Our semantic taxonomy is tailored to sidewalk-centric urban scenes.
We define 15 semantic groups and 39 member text prompts for 2D pseudo-labeling (Table~\ref{tab:taxonomy}).
Among them, 14 groups are lifted to 3D for semantic occupancy supervision; \texttt{occlusion} is a 2D-only label used to mark image regions without valid scene semantics (e.g., fisheye borders) and is excluded during 3D projection.
The 14 occupancy classes are {\textcolor{background}{$\blacksquare$} background, {\textcolor{road}{$\blacksquare$} road, {\textcolor{sidewalk}{$\blacksquare$} sidewalk, {\textcolor{crosswalk}{$\blacksquare$} crosswalk, {\textcolor{grass}{$\blacksquare$} grass, {\textcolor{wall}{$\blacksquare$} wall (building and fence), {\textcolor{trees}{$\blacksquare$} trees, {\textcolor{vehicle}{$\blacksquare$} vehicle, {\textcolor{cyclist}{$\blacksquare$} cyclist (including scooter and wheelchair), {\textcolor{animal}{$\blacksquare$} animal, {\textcolor{generic_obstacle}{$\blacksquare$} generic obstacle (for example, pillar, bench, fire hydrant, stroller), {\textcolor{pedestrian}{$\blacksquare$} pedestrian, {\textcolor{curb}{$\blacksquare$} curb, and {\textcolor{gutter}{$\blacksquare$} gutter; the full 2D prompt list, including occlusion, appears in Table~\ref{tab:taxonomy}.

\begin{table}[t]
  \centering
  \caption{Semantic taxonomy: 15 groups and 39 member prompts for 2D labeling. Occlusion is 2D-only and is not included in the 14-class 3D occupancy taxonomy.}
  \label{tab:taxonomy}
  \small
  \begin{tabular}{@{}clp{6.5cm}@{}}
    \toprule
    ID & Group & Member prompts \\
    \midrule
    0 & background & \texttt{background} \\
    1 & road & \texttt{road} \\
    2 & sidewalk & \texttt{sidewalk} \\
    3 & crosswalk & \texttt{crosswalk} \\
    4 & grass & \texttt{natural}, \texttt{lawn} \\
    5 & wall & \texttt{wall}, \texttt{building}, \texttt{fence} \\
    6 & trees & \texttt{trees}, \texttt{potted plant}, \texttt{planter}, \texttt{flower pot} \\
    7 & vehicle & \texttt{vehicle} \\
    8 & cyclist & \texttt{cyclist}, \texttt{scooter}, \texttt{wheelchair} \\
    9 & animal & \texttt{animal} \\
    10 & generic obstacle & \texttt{pillar}, \texttt{signboard}, \texttt{chair}, \texttt{desk}, \texttt{table}, \texttt{stool}, \texttt{bench}, \texttt{sofa}, \texttt{bed}, \texttt{cabinet}, \texttt{shelf}, \texttt{drawer}, \texttt{fire hydrant}, \texttt{stroller}, \texttt{bike rack} \\
    11 & pedestrian & \texttt{person}, \texttt{rider} \\
    12 & curb & \texttt{curb} \\
    13 & gutter & \texttt{gutter} \\
    14 & occlusion (2D only) & \texttt{occlusion} \\
    \bottomrule
  \end{tabular}
\end{table}

\begin{algorithm}[t]
  \DontPrintSemicolon
  \KwInput{Front-view images $\{I_t\}$, LiDAR points $\{P_t\}$, calibration $\{\mathbf{T}_{\mathrm{lidar}\to\mathrm{cam}},\mathbf{K}\}$, SAM3 checkpoints $\theta_{\mathrm{base}},\theta_{\mathrm{ft}}$, prompt set $\Pi$, (optional) ego poses $\{\mathbf{T}_{t}\}$}
  \KwOutput{Semantic occupancy ground truth $\mathcal{O}$}
  \BlankLine
  \textbf{Stage 1: 2D pseudo-labels (overcomplete prompts)}\;\tcp*[r]{per frame (front view)}
  \For{each frame $t$}{
    $L^{\mathrm{base}}_t \leftarrow \text{MergePrompts}(\text{SAM3}(I_t,\theta_{\mathrm{base}},\Pi))$\;
    $L^{\mathrm{ft}}_t   \leftarrow \text{MergePrompts}(\text{SAM3}(I_t,\theta_{\mathrm{ft}},\Pi))$\;\tcp*[r]{priority overwrites}
    $L^{2\mathrm{D}}_t \leftarrow \text{ClassAwareMerge}(L^{\mathrm{base}}_t, L^{\mathrm{ft}}_t)$\;
  }
  \BlankLine
  \textbf{Stage 2: 3D lift / point cloud colorization}\;\tcp*[r]{project LiDAR points to the front-view image plane}
  \For{each frame $t$}{
    \For{each point $\mathbf{p}\in P_t$}{
      Find a valid projection $(u,v)$ using $\mathbf{T}_{\mathrm{lidar}\to\mathrm{cam}}$ and $\mathbf{K}$\;
      \If{valid and $(u,v)$ in image}{assign $\ell(\mathbf{p})\leftarrow L^{2\mathrm{D}}_t(u,v)$\;}
    }
  }
  \BlankLine
  \textbf{Stage 3 (optional): Voting to suppress semantic flicker}\;\tcp*[r]{voxel/KNN majority vote}
  $P_{\mathrm{all}} \leftarrow \text{AccumulateInWorld}(\{P_t\},\{\mathbf{T}_t\})$\;
  $P_{\mathrm{all}} \leftarrow \text{MajorityVote}(P_{\mathrm{all}})$\;
  $\{P_t\} \leftarrow \text{SplitBack}(P_{\mathrm{all}},\{\mathbf{T}_t\})$\;
  \BlankLine
  \textbf{Stage 4: Occupancy GT generation}\;\tcp*[r]{SurroundOcc preprocessing}
  $\mathcal{O} \leftarrow \text{SurroundOccPreprocess}(\{P_t\},\text{calib},\text{poses},\text{boxes})$\;
  \caption{Pseudo-label pipeline.}
  \label{fig:pseudo-pipeline-pseudocode}
\end{algorithm}

\subsection{Overcomplete Prompts and Label Merging}
\label{sec:overcomplete-prompts}

SAM3 accepts only one text prompt per inference. To obtain dense per-pixel labels over our taxonomy, we use an \emph{overcomplete prompt set}: 39 fine-grained prompts covering all 2D member classes (Table~\ref{tab:taxonomy}). Since prompts are processed independently, we run SAM3 multiple times with different prompt subsets and then merge the resulting masks.

When masks overlap, we apply a deterministic overwrite rule. For the finetuned pass, we use a low-to-high priority order
\texttt{sidewalk} $\to$ \texttt{road} $\to$ \texttt{driveway} $\to$ \texttt{crosswalk} $\to$ \texttt{gutter} $\to$ \texttt{natural} $\to$ \texttt{curb} $\to$ \texttt{vehicle} $\to$ \texttt{rider} $\to$ \texttt{person} $\to$ \texttt{animal} $\to$ \texttt{occlusion},
so later classes overwrite earlier ones in overlapping pixels. For the base pass, we use no explicit priority and simply overwrite in prompt order. Example splits:
\begin{itemize}
  \item \textbf{{Priority layer}} (finetuned pass): \texttt{sidewalk}, \texttt{road}, \texttt{crosswalk}, \texttt{curb}, \texttt{gutter}, \texttt{driveway}, \texttt{natural}, \texttt{vehicle}, \texttt{rider}, \texttt{person}, \texttt{animal}, \texttt{occlusion}
  \item \textbf{Complement} (base pass): \texttt{lawn}, \texttt{wall}, \texttt{building}, \texttt{fence}, \texttt{trees}, \texttt{cyclist}, \texttt{scooter}, \texttt{wheelchair}, and generic obstacles (\texttt{chair}, \texttt{desk}, \texttt{bench}, \texttt{fire hydrant}, \texttt{signboard}, etc.)
\end{itemize}
For each frame, we run both a pretrained base checkpoint and a checkpoint finetuned on our data, producing $L^{\mathrm{base}}$ and $L^{\mathrm{ft}}$.
We start from $L=L^{\mathrm{base}}$ and overwrite with finetuned predictions where available, except for the cyclist group where the base model is more reliable: $L[\mathcal{M}] = L^{\mathrm{ft}}[\mathcal{M}]$ with $\mathcal{M} = (L^{\mathrm{ft}} \neq 0) \wedge (L^{\mathrm{base}} \notin \mathcal{C}_{\mathrm{cyclist}})$.
The merged mask $L$ is used as our 2D pseudo-label for lifting to 3D.

\subsection{3D Lift and Point Cloud Colorization}
\label{sec:3d-lift}
We lift 2D pseudo-labels to 3D by transforming each LiDAR point to the camera frame with $\mathbf{T}_{\mathrm{lidar}\to\mathrm{cam}}$ and projecting it onto the image plane using the calibrated camera model (pinhole or fisheye) and intrinsics $\mathbf{K}$.
If the projected pixel $(u,v)$ lies inside the image, the projection is valid, and the 2D label is not occlusion, we assign the point the semantic label at $(u,v)$ from the merged 2D mask; otherwise the point is left unlabeled.
This produces a per-frame colored point cloud with 14 semantic classes.

\subsection{Voting to Suppress Semantic Flicker}
\label{sec:voting-flicker}
Because each 3D point is labeled from only one camera view in a single frame, projection noise, occlusions, and temporal inconsistencies in the 2D masks may assign different semantic labels to the same physical region across frames or nearby points (semantic flicker).
To mitigate this, we insert an optional voting step before SurroundOcc conversion.
We merge all frames in a sequence into a single global point cloud in world coordinates using per-frame poses $\mathbf{T}_{\mathrm{lidar}\to\mathrm{world}}$, voxelize it with a fixed voxel size (e.g., $0.5\,\mathrm{m}$), and perform majority voting over the semantic labels inside each voxel (ignoring background).
All points in a voxel are reassigned to the winning class.
We then split the voted global cloud by frame index, transform each frame back to its LiDAR coordinate system, and overwrite the per-frame colored point clouds with the voted labels.
As an alternative, we implement a KNN-based variant that, for each point, queries its $k$ nearest neighbors with a KDTree and replaces the label with the majority class if it exceeds a threshold ratio (default 0.6).
The voted point clouds are then fed into the SurroundOcc conversion pipeline.

\subsection{SurroundOcc Format and Occupancy GT}
\label{sec:surroundocc-gt}
Starting from the (optionally voted) per-frame semantic point clouds, we convert each sequence to the data layout expected by SurroundOcc and follow its official preprocessing protocol~\cite{wei2023surroundocc}.
Dynamic points are associated across time using box identities/tokens, whereas static structure is accumulated in the world frame.
For each region, multi-frame points are fused, a surface mesh is reconstructed via Poisson reconstruction, and the mesh is voxelized into a 14-class semantic occupancy grid at a user-defined resolution and 3D range.
The resulting labels are provided over a fixed temporal window, and preprocessing can be parallelized across scenes for scalability.

\section{Sidewalk-Specialized SAM3 Fine-tuning}\label{app:sam3}

We fine-tuned SAM3 \cite{carion2025sam} on our robot-collected sidewalk dataset. The fine-tuned model is used to generate 2D segmentation masks for the 3D reasoning pipeline in this paper.

Fine-tuning is necessary for two reasons. First, our visual domain differs from the data used in SAM3 pretraining. Images are captured from a ground-level mobile robot with a fisheye camera, which introduces viewpoint bias, geometric distortion, and a circular lens boundary. The scenes are sidewalk-centered rather than object-centric. These factors lead to degraded performance when the pretrained model is applied directly.

Second, our labels differ from those used in general segmentation benchmarks. In addition to standard classes such as pedestrian and vehicle, our task uses a sidewalk-specific taxonomy that includes curb, crosswalk, and gutter, each of which is defined within our task specification. These categories are not consistently defined or separated in existing datasets. Fine-tuning allows the model to align with our task-specific taxonomy.

\subsection{2D Segmentation Dataset}

We sample 6,000 images from a larger corpus collected by a ground-level mobile robot equipped with a fisheye camera. Due to the lens geometry, each image contains a circular black border, which is annotated as an ``occlusion'' class. In addition, the dataset includes nine semantic classes: pedestrian, vehicle, animal, curb, crosswalk, grass, sidewalk, road, and gutter. Annotations are produced on Segments.ai by around 20 annotators over approximately one month.

After filtering low-quality images (e.g., strong blur or severe obstruction), 5,626 images remain. The dataset is split into 4,830 training images and 796 validation images. The split is performed at the robot-unit level, so data collected by the same unit appears in only one split, reducing leakage from repeated routes and visually similar scenes.

\subsection{Implementation Details}
\begin{figure}[htbp]
    \centering
    \includegraphics[width=\textwidth]{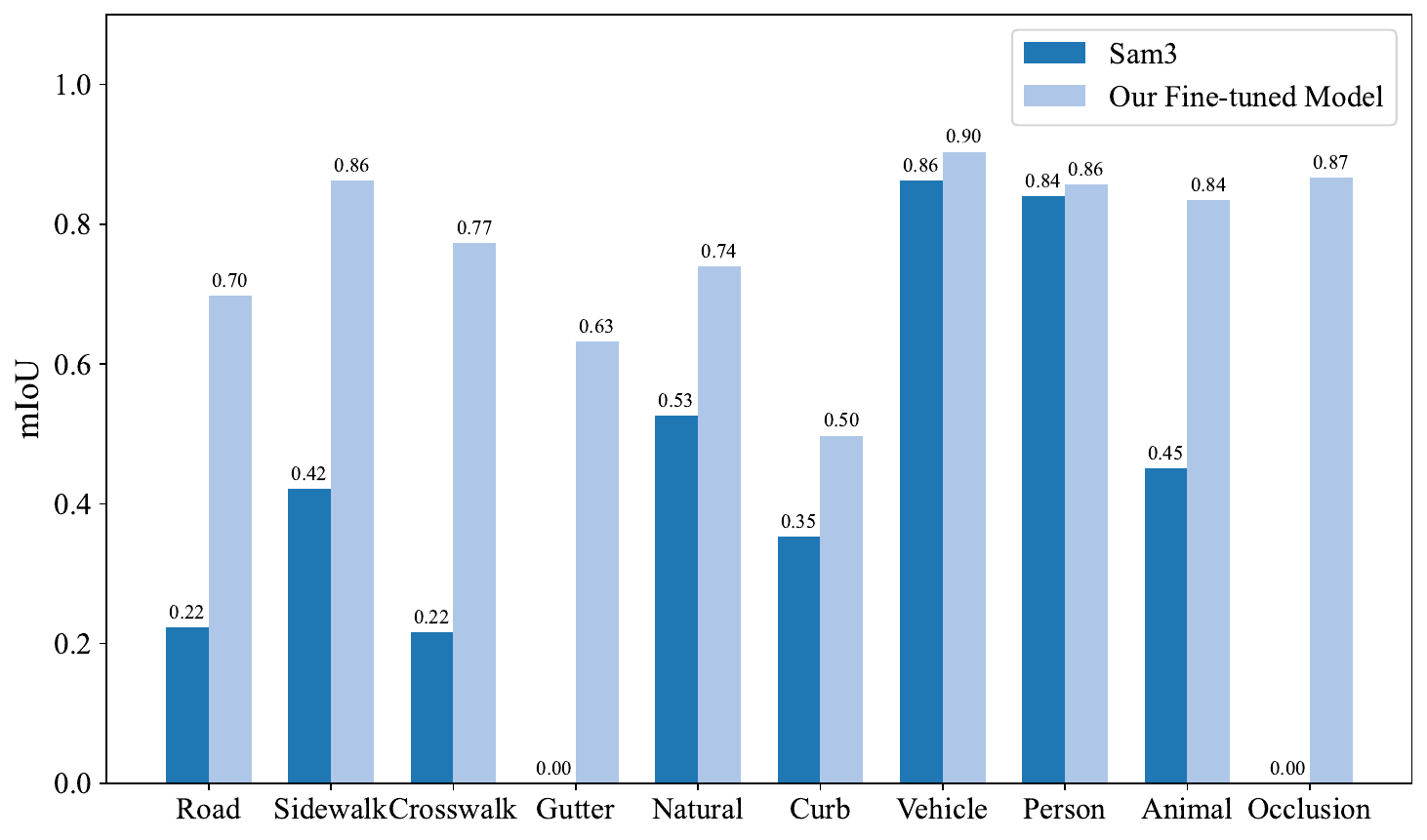}
    \caption{Per-class macro-averaged mIoU on the validation set for pretrained SAM3 and our fine-tuned model. Scores are computed on merged semantic maps after mask composition.}
    \label{fig:miou_compare}
\end{figure}
We follow the official \texttt{roboflow\_v100} training configuration and adapt it to our COCO-format dataset. Segmentation training is enabled, while the original augmentation and preprocessing pipeline is retained, including random box perturbation, resizing and padding to 1008, and normalization with mean/std = 0.5. COCO masks are decoded in both training and validation.

The model is initialized from the released SAM3 pretrained weights. We freeze the language backbone (learning rate = 0) and train the vision backbone with a constant learning rate of $1.6\times10^{-6}$. The remaining modules are optimized using AdamW with an inverse-square-root schedule, with a base learning rate of $8\times10^{-5}$. Training runs for 100 epochs using bfloat16 mixed precision and gradient clipping (max norm = 0.1). Experiments are conducted on 8 GPUs with a batch size of 2 for training and 4 for validation, with validation performed every 5 epochs.

Checkpoint selection is based on validation performance. We monitor cgF1$_{\text{segm}}$ and macro mIoU during training. While cgF1 remains stable from epochs 45--80, macro mIoU peaks at epoch 50. The epoch-50 checkpoint is used in subsequent experiments.

\subsection{Qualitative Visualization}
\begin{figure}[htbp]
    \centering
    \includegraphics[width=0.88\textwidth]{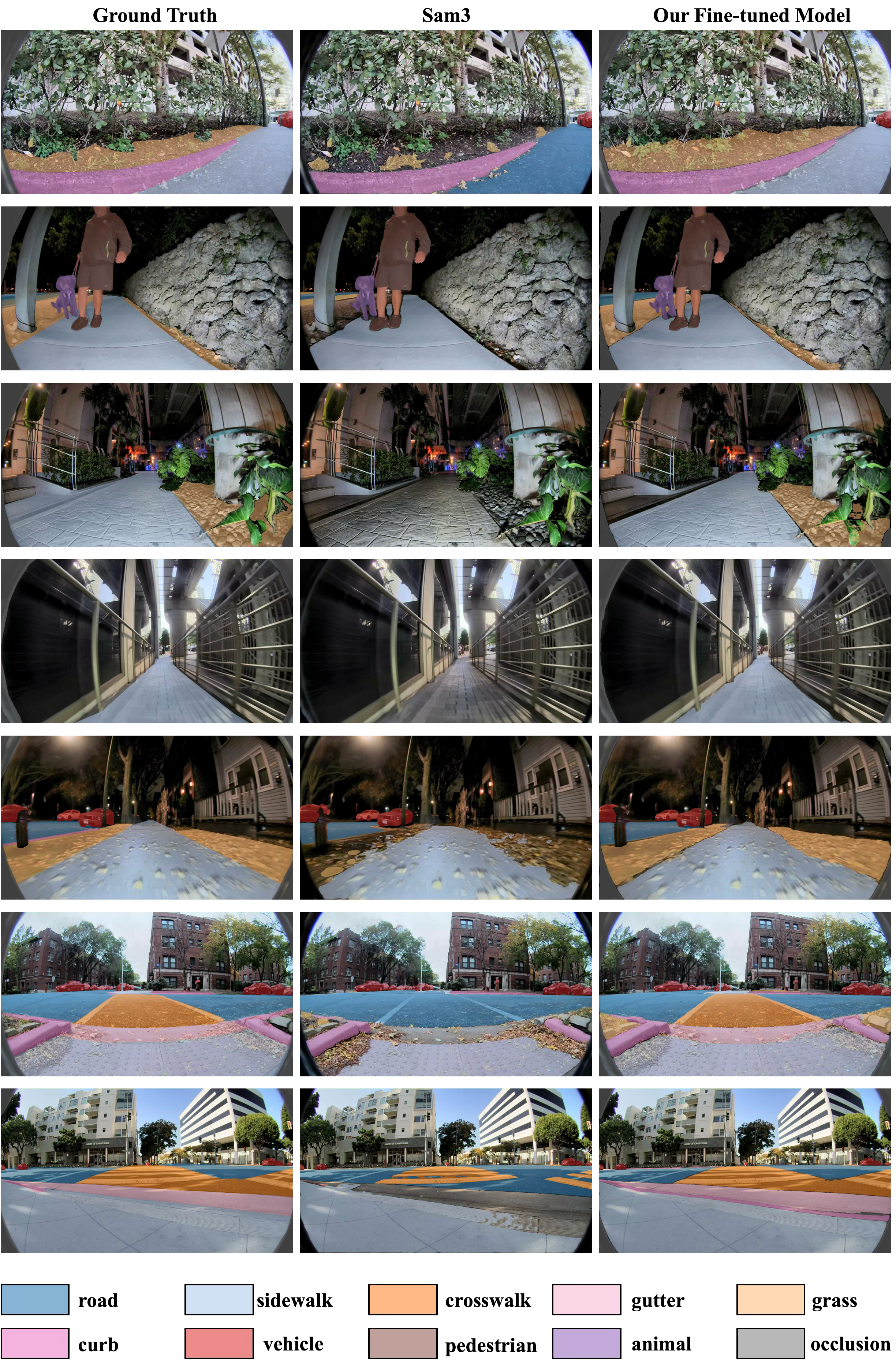}
    \caption{Qualitative comparison on the validation set. From left to right: ground truth, pretrained SAM3, and our fine-tuned model.}
    \label{fig:example}
\end{figure}
Segmentation is generated independently for each class using class-specific prompts, and the resulting masks are merged into a single semantic map. Overlapping regions are resolved using a fixed priority order: road and sidewalk at low priority; crosswalk, gutter, grass, and curb at intermediate priority; and vehicle, pedestrian, and animal at higher priority. The occlusion mask is assigned the highest priority.

All quantitative results are computed on the merged semantic maps. Figure~\ref{fig:miou_compare} reports per-class macro mIoU on the validation set for the pretrained SAM3 baseline and our fine-tuned model. For common object prompts such as vehicle and pedestrian, pretrained SAM3 already achieves relatively high mIoU. In contrast, clear improvements are observed after fine-tuning for ground-related classes, including road, sidewalk, crosswalk, and curb. For the gutter, which lies at the boundary between the road and the curb, the pretrained model fails to produce meaningful predictions.

Figure~\ref{fig:example} presents qualitative results on the validation set. Columns correspond to ground truth, the official pretrained SAM3 model, and our fine-tuned model. The fine-tuned model reduces confusion between adjacent ground classes (sidewalk, road, crosswalk) and improves boundary delineation for curb and gutter. Improvements are also visible near the fisheye boundary and the occlusion region.

For grass surfaces, Figure~\ref{fig:example} also illustrates a limitation of the pretrained baseline. It is very likely that in our task, the definition of grass regions includes non-paved ground beyond simply greenery (e.g., exposed soil or sparse vegetation). As a result, the baseline may miss grass areas that are not visually grass-like.

\section{Dataset Collection and Preprocessing Details}\label{app:data-preprocess}


\subsection{{\DatasetName} for Urban Perception}

\begin{figure}[t]
    \centering
    \includegraphics[width=\linewidth]{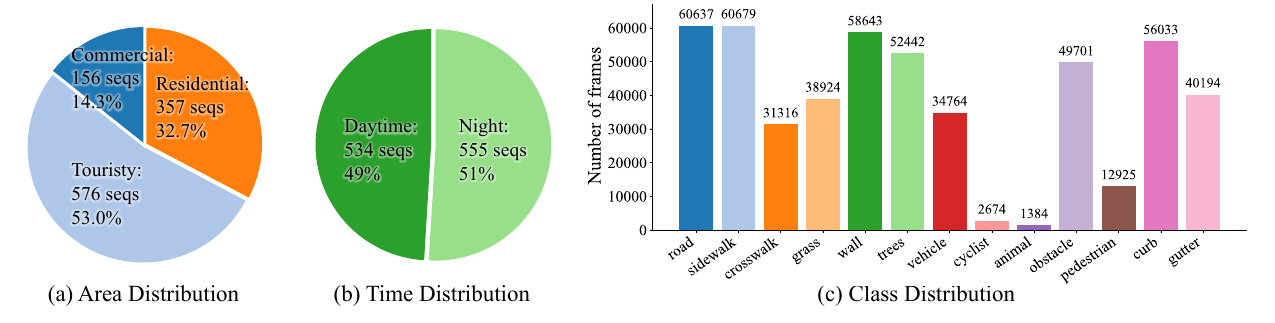}
    \caption{\textbf{Data distribution and sample scenes from {\DatasetName}}. Our dataset spans diverse domains, regions, and illumination conditions (day/night). 
    }
    \label{fig:datadistrubution}
\end{figure}
To better train and evaluate our method, we construct a large-scale perception dataset, \DatasetName, tailored to micro-mobility in real-world urban environments.
As shown in Figure~\ref{fig:datadistrubution}(a), the data are collected from three area types: \textbf{touristy} (576 sequences, 53\%), \textbf{residential} (357 sequences, 32.7\%), and \textbf{commercial} (156 sequences, 14.3\%) districts. Each sequence contains 60 synchronized LiDAR--camera frames captured at 2\,Hz, together with metric poses. The dataset spans diverse illumination conditions, with 555 sequences (51\%) recorded at night and the remainder during the day (Figure~\ref{fig:datadistrubution}(b)). The class distribution is summarized in Figure~\ref{fig:datadistrubution}(c); most classes are relatively balanced, while cyclists and animals are under-represented. Representative scenes include scenic viewpoints with dense vegetation and many shops (touristy), neighborhoods with abundant grass and houses (residential), and downtown regions with structured roads and dense buildings (commercial). Nighttime sequences are particularly challenging due to low-light conditions and motion-induced trailing (ghosting). Overall, this cross-domain, multi-condition dataset provides a valuable testbed for robust micro-mobility perception.

We split the dataset into 928 sequences for training and 103 sequences for validation.
For evaluation, we manually refine pseudo-labels in the test scenarios to obtain high-quality occupancy ground truth. Specifically, we refine 18 touristy-nighttime sequences, 16 sequences from commercial districts, and 23 additional touristy-daytime sequences. We first refine the point-cloud semantic segmentation results and then generate occupancy labels accordingly. Figure~\ref{fig:dataset} shows refined examples, including zoomed-in views with the corresponding RGB images, LiDAR semantic points, and the resulting occupancy ground truth.

\begin{figure}[t]
    \centering
    \includegraphics[width=\linewidth]{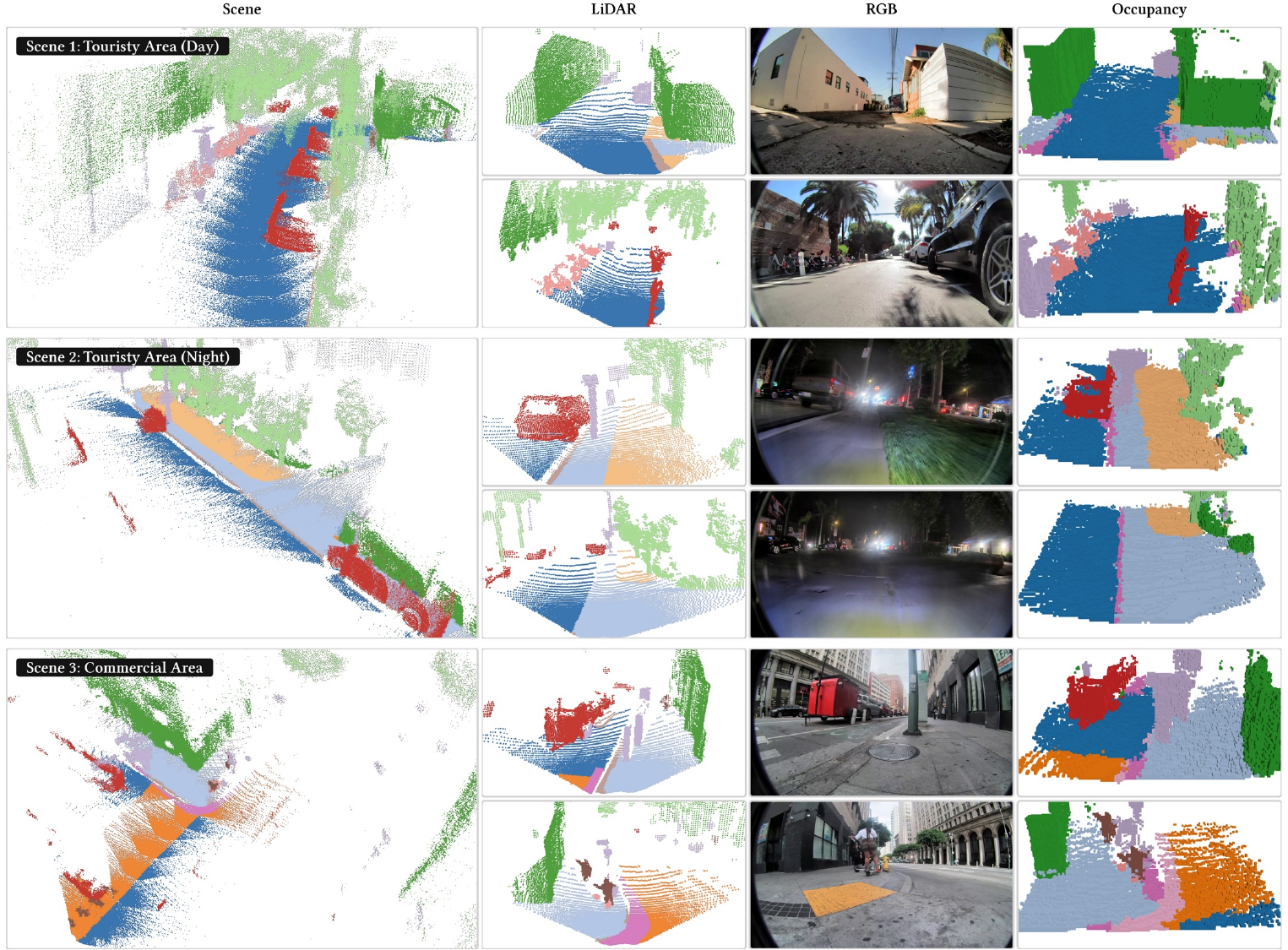}
    \caption{\textbf{Refined occupancy ground truth examples.} We visualize the manually annotated global point clouds for three representative scenarios: touristy--day, touristy--night, and commercial. For each scenario, the right panel shows a nearby sample with its LiDAR semantic point cloud, the corresponding RGB image, and the refined occupancy ground truth. This figure highlights the unstructured nature and diversity of real-world urban-robot scenes, as well as the large discrepancies across locations and illumination conditions, underscoring the importance of our framework and dataset.}
    \label{fig:dataset}
\end{figure}

\subsection{Data Collection}
We collected $10$ hours of data from $3$ robots driving across different regions. Each robot is equipped with a front camera capturing visual frames at $20$ fps, a lidar which captures 3D scene geometry at $10$ fps, and an imu that records linear acceleration and angular velocity at $40$ fps. To ensure data diversity, we picked visually distinct regions, namely tourist, residential, and business areas.

\subsection{Data Preprocessing Pipeline}
From the raw data collected, we performed some basic filtering of clips where the robot was static or in an invalid state (e.g., flipped over), in order to remove low-quality scenarios.

\begin{wrapfigure}{l}{0.5\columnwidth}
    \vspace{-2pt}
    \includegraphics[width=\linewidth]{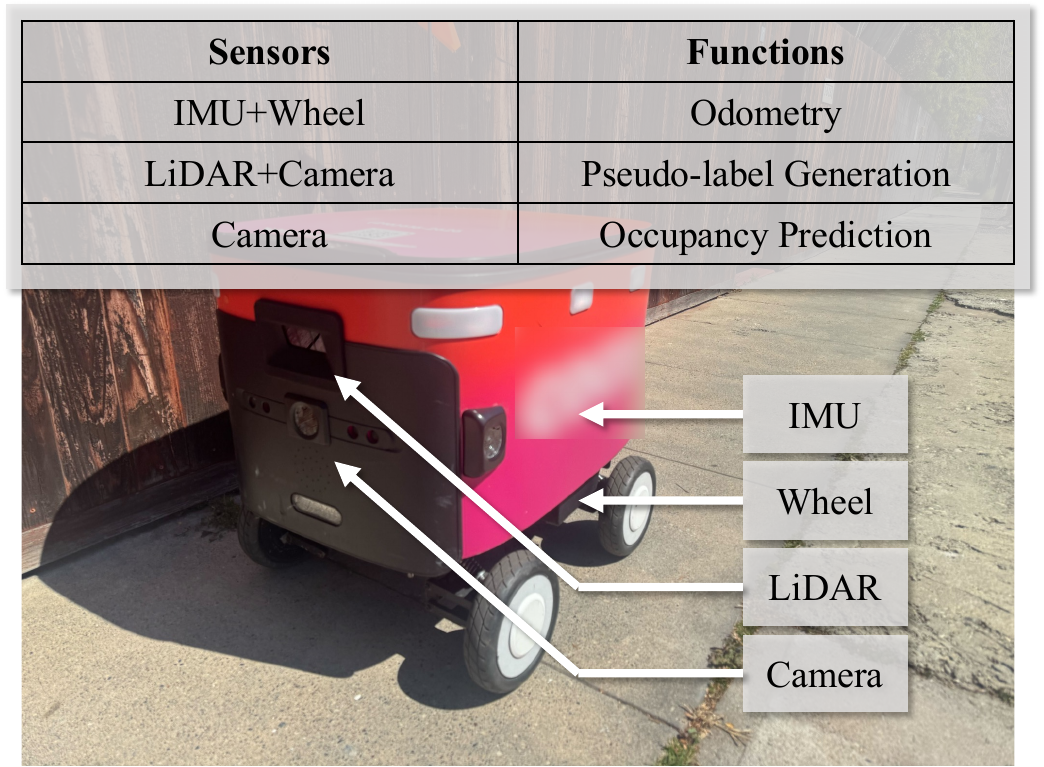}
    \caption{
    \textbf{Data collection platform.} Our mobile robot is equipped with a forward-facing RGB camera, a 3D LiDAR, an IMU, and wheel odometry for pose estimation.
    }
    \label{fig:platform}
    \vspace{-20pt}
\end{wrapfigure}
All post-filtered data is split into $30$ second scenes at $1$ fps. Each scene is fed into an EKF-based algorithm that uses IMU measurements and on-board wheel odometry to recover the robot's pose.

With the robot odometry, we can then create an aggregated scene pointcloud, which can be reprojected onto each camera frame. Using our pseudo-label generation process as detailed in Section 3.3, we obtain initial per-frame segmentation masks.

\noindent\textbf{Sensor Calibration}
We calibrated the front camera and lidar with standard calibration methods to obtain the camera extrinsics and intrinsics. A Charuco board with a reflective tape ring was used for this process.

\noindent\textbf{Time Synchronization}
Because the camera and imu sensors are not time synchronized, we perform a least squares solve of the time offset by comparing SfM camera poses with IMU-derived poses. We assume the time offset between imu measurements and lidar reads is negligible.

\noindent\textbf{Manual Refinement of Pseudo Labels}
We used an annotation platform to refine the 3D semantic pointcloud pseudolabels, which primarily involved referring to the scene geometry for more smooth and precise labels. Dynamic objects were also carefully tuned in per-frame 3D views.

\end{document}